\documentclass[10pt,twocolumn,letterpaper]{article}

\usepackage{cvpr}              

\usepackage[utf8]{inputenc} 
\usepackage[T1]{fontenc}    
\usepackage{url}            
\usepackage{booktabs}       
\usepackage{amsfonts}       
\usepackage{nicefrac}       
\usepackage{microtype}      
\usepackage{xcolor}         
\usepackage{algpseudocode}
\usepackage[ruled,linesnumbered]{algorithm2e}
\usepackage{bm}
\usepackage{amsmath}
\usepackage{bbm}
\usepackage{bbding}

\usepackage{float}
\usepackage{multirow}

\usepackage{graphicx}
\usepackage{subcaption}
\usepackage{amsthm}
\usepackage{thmtools, thm-restate}
\usepackage{amssymb}
\usepackage{comment}
\usepackage[numbers]{natbib}
\usepackage{tabularx}
\usepackage[cjk]{kotex}
\usepackage[inkscapelatex=false]{svg}

\allowdisplaybreaks

\newtheorem{lemma}{Lemma}

\newtheorem{definition}{Definition}
\newtheorem{proposition}{Proposition}

\theoremstyle{remark}

\DeclareMathOperator*{\argmin}{argmin}
\DeclareMathOperator*{\argmax}{argmax}

\newcommand{\advmax}{{\underset{\mathbf{X}'\in\mathcal{B}_p(\mathbf{X},\varepsilon)}{\max}}}

\newcommand{\expD}{{\mathbb{E}_{\widehat{\mathcal{D}},\sigma}}}
\newcommand{\cm}{{\mathcal{M}}}

\definecolor{cvprblue}{rgb}{0.21,0.49,0.74}
\usepackage[pagebackref,breaklinks,colorlinks,allcolors=cvprblue]{hyperref}

\def\paperID{13368} 
\def\confName{CVPR}
\def\confYear{2025}

\title{TAROT: Towards Essentially Domain-Invariant Robustness\\with Theoretical Justification}

\author{
Dongyoon Yang$^{1*\dagger}$\quad Jihu Lee$^{2*}$\quad Yongdai Kim$^{2\dagger}$\\
$^1$AI Advanced Technology, SK Hynix\\
$^2$Department of Statistics, Seoul National University\\
{\tt\small ydy0415@gmail.com, rieky0426@snu.ac.kr, ydkim903@snu.ac.kr}
}

\begin{document}
\maketitle
\def\thefootnote{*}\footnotetext{Equal contribution.}
\def\thefootnote{$\dagger$}\footnotetext{Corresponding authors.}
\begin{abstract}
Robust domain adaptation against adversarial attacks is a critical research area that aims to develop models capable of maintaining consistent performance across diverse and challenging domains. In this paper, we derive a new generalization bound for robust risk on the target domain using a novel divergence measure specifically designed for robust domain adaptation.
Building upon this, we propose a new algorithm named TAROT, which is designed to enhance both domain adaptability and robustness. 
Through extensive experiments, TAROT not only surpasses state-of-the-art methods in accuracy and robustness but also significantly enhances domain generalization and scalability by effectively learning domain-invariant features. In particular, TAROT achieves superior performance on the challenging DomainNet dataset, demonstrating its ability to learn domain-invariant representations that generalize well across different domains, including unseen ones.
These results highlight the broader applicability of our approach in real-world domain adaptation scenarios.
\vskip -0.2in
\end{abstract}

\section{Introduction}
\label{sec:intro}

Recent advances in machine learning have highlighted the importance of developing transferable models that can maintain performance across different domains. However, collecting labeled data from all potential domains is too resource-intensive for many practical applications. To address this limitation, Unsupervised Domain Adaptation (UDA) has emerged as a type of transfer learning that leverages labeled source-domain data to improve performance on a target domain without needing labels from that domain. UDA has proven pivotal in diverse machine learning tasks, largely because distributional discrepancies between source and target domains often lead to severe performance degradation \cite{johansson2019support, munir2021sstn, barbato2021latent, singhal2023domain, teichmann2018multinet}.

Meanwhile, adversarial robustness has gained significant attention as a crucial element of secure and trustworthy AI. 
Pioneering works
demonstrated the vulnerability of neural networks to small, carefully designed perturbations called \textit{adversarial attacks} \cite{szegedy2014intriguing, madry2018towards}
and proposed adversarial training methods
\cite{madry2018towards, zhang2019theoretically, wang2020improving, yang2023improving, yang2023enhancing}.
Recently, the adversarial robustness community has expanded its focus to multi-modal tasks (e.g., text-to-image, image-to-text), recognizing that large-scale multi-modal models are also susceptible to adversarial manipulations in one or more input modalities \cite{kim2024race, gong2024reliable, schlarmann2024robust, mao2023understanding}. 
These cross-modal attacks can exploit alignment mechanisms to trick models into misclassification or erroneous content generation.

Given these two developments--domain adaptation and adversarial robustness--there is a growing interest in learning transferable models in UDA settings that remain resilient to adversarial perturbations in the target domain \cite{awais2021adversarial, gao2022certifying, lo2022exploring, zhu2023srouda}. 
A common strategy in existing robust UDA methods is to rely on pseudo-labels to handle unlabeled target data. However, in this paper, we propose a novel robust margin disparity metric for comparing source and target distributions, enabling us to derive a refined generalization bound on the robust risk in the target domain. From our bound, we derive two key insights: (1) using only pseudo-labels may be insufficient, but combining pseudo-labels with distributional alignment significantly enhances robust UDA performance; and (2) initializing robust UDA algorithms with a robustly pre-trained model (Robust-PT \cite{awais2021adversarial}) is notably more effective than starting from a non-robust pre-trained model.

Building upon these theoretical foundations, we propose TAROT (Transfer Adversarially RObust Training), a new algorithm that leverages our theoretical insights to achieve superior performance in robust UDA tasks. 
Through extensive experiments on various benchmark datasets, we demonstrate that
TAROT outperforms existing algorithms. 
Furthermore, we empirically demonstrate that TAROT exhibits distinct advantages in terms of scalability compared to pseudo-labeling approaches, particularly with significant performance improvements not only on the target domain but also on source and unseen source domains.
This work bridges the gap between theoretical understanding and practical implementation in robust transfer learning, offering both fundamental insights and actionable solutions for real-world applications. 

Our contributions are summarized as follows:
\begin{itemize}
    \item We introduce a novel robust margin disparity measure and derive a refined generalization bound for robust UDA.
    \item From the derived generalization bound, we propose TAROT (\textbf{T}ransferring \textbf{A}dversarially \textbf{RO}bust \textbf{T}raining), which integrates pseudo-labeling and distributional alignment to enhance adversarial robustness in UDA.
    \item TAROT outperforms existing robust UDA methods across multiple benchmarks, demonstrating strong performance on the target, source, and unseen source domains, indicating that TAROT effectively facilitates domain-invariant robust representations.
    \item We provide theoretical insights and practical solutions for scalable and adversarially robust domain adaptation in real-world scenarios.
\end{itemize}
\section{Preliminaries}
\label{sec:preliminaries}
In this section, we outline the core concepts and notations for our analysis, beginning with adversarial training and its risk formulation for robust learning. We then describe the unsupervised domain adaptation (UDA) framework, highlighting distributional notations, the formal UDA objective, and existing theoretical approaches like DANN \cite{ganin2016domain} and MDD \cite{zhang2019bridging}. These elements provide the groundwork for our subsequent derivation of a robust UDA upper bound and the proposed learning algorithm.

\subsection{Adversarial Training}
Let $\mathcal{X} \in \mathbb{R}^d$ represent the input space,
and let $\mathcal{Y} = \{1, \dots, C\}$ be the label space, where $C$ represents the number of classes.
A classifier can be represented using a \textit{score function} $f: \mathcal{X} \to \mathbb{R}^C$, where $f_c$ denotes the $c$-th component of $f$. 
The classifier is then defined as:  
$
h_f(\bm{x}) = \underset{c \in \mathcal{Y}}{\argmax} f_c(\bm{x}).
$
Let $\mathcal{B}_p(\bm{x}, \varepsilon) = \{\bm{x}' \in \mathcal{X} : \lVert \bm{x} - \bm{x}' \rVert_p \leq \varepsilon\}$ denote an \textit{$\varepsilon$-ball centered at} $\bm{x}$.

The population risk considering adversarial robustness is defined as follows: 
\begin{equation}
    \label{eq:rob_risk}
    \mathcal{R}_{\text{rob}}(f)=\mathbb{E}_{(\mathbf{X},Y)} \;  \underset{\mathbf{X}' \in \mathcal{B}_p(\mathbf{X}, \varepsilon) }{\max\;\;\;} \mathbbm{1}\left\{Y\ne h_f(\mathbf{X'}) \right\},
\end{equation}
where $\mathbbm{1}(\cdot)$ denotes the indicator function.
The goal of adversarial training is to estimate the optimal $f$, the minimizer of \cref{eq:rob_risk}. 

\subsection{Unsupervised Domain Adaptation}

There are two distributions, the source distribution $\mathcal{S}$ and the target distribution $\mathcal{T}$. The empirical versions of $\mathcal{S}$ and $\mathcal{T}$ are $\widehat{\mathcal{S}} = \{(\bm{x}^s_i, y^s_i)\}_{i=1}^{n_s}$ and $\widehat{\mathcal{T}} = \{(\bm{x}^t_j, y^t_j)\}_{j=1}^{n_t}$, respectively. In unsupervised domain adaptation, $\widehat{\mathcal{S}}$ and only $\widehat{\mathcal{T}}_{\mathbf{X}}:= \{\bm{x}^t_j\}_{j=1}^{n_t}$ are observed. i.e. the labels $y^t_j$ can not be observed. Denote the distributions of the input on the source and target domains by $\mathcal{S}_{{\mathbf{X}}}$ and $\mathcal{T}_{\mathbf{X}},$ respectively.

For a given score function $f \in \mathcal{F}$, the risk of a data distribution $\mathcal{D}$ is defined as
\begin{equation}
\mathcal{R}_{\mathcal{D}}(f):= \mathbb{E}_{(\mathbf{X}, Y) \sim \mathcal{D}} 
 \mathbbm{1}\{ Y\ne h_f(\mathbf{X}) \}.
\end{equation}
The goal of UDA is to minimize $\mathcal{R}_{\mathcal{T}}(f)$ using $\widehat{\mathcal{S}}$ and $\widehat{\mathcal{T}}_{\mathbf{X}}$, which can be done by building an upper bound of $\mathcal{R}_{\mathcal{T}}(h)$.

The pivotal learning algorithms for UDA are DANN (Domain-Adversarial Neural Network) \cite{ganin2016domain} and MDD (Margin Disparity Discrepancy) \cite{zhang2019bridging}. 
\citet{ben2010theory} suggested a distance $\mathit{d}_{\mathcal{H}\Delta \mathcal{H}} (\mathcal{S}_\mathbf{X}, \mathcal{T}_\mathbf{X})$ between $\mathcal{S}_\mathbf{X}$ and $\mathcal{T}_\mathbf{X}$ to give an upper bound
of the robust risk, and \citet{ganin2016domain} utilized this upper bound to implement  their learning algorithm
DANN.  DANN uses an adversarial neural network to minimize the distance between the source and target
distributions, using gradient reversal layers (GRL). 
On the other hand, MDD \cite{zhang2019bridging} derives an upper bound of $\mathcal{R}_{\mathcal{T}}(f)$ as the sum of $\mathcal{R}^{(\rho)}_{\mathcal{S}}(f)$, a standard non-robust risk on the source domain, and $\mathit{d}^{(\rho)}_{f, \mathcal{F}}(\mathcal{S}_\mathbf{X}, \mathcal{T}_\mathbf{X})$, a distance between $\mathcal{S}_\mathbf{X}$ and $\mathcal{T}_\mathbf{X}$,
and develop the MDD algorithm based on this upper bound.
The main distinction between the two upper bounds is that MDD utilizes a scoring function for multi-class classification, while DANN only employs a classifier for binary classification.

The objective of our work is to derive an upper bound of the robust risk for the target domain by utilizing the risk in the source domain and the robust divergence between the target domain $\mathcal{T}$ and the source domain $\mathcal{S}$. In turn, inspired by the upper bound, we propose a novel robust UDA algorithm. 
\section{Related Works}
\label{sec:related_works}

While a significant body of research on adversarial training has focused on supervised learning and semi-supervised learning tasks \cite{madry2018towards, zhang2019theoretically, yang2023enhancing, yang2023improving, uesato2019are}, there has been relatively less attention given to its application in domain adaptation tasks . Only a few studies have explored adversarial training in the context of domain adaptation \cite{awais2021adversarial, zhu2023srouda, lo2022exploring}, highlighting the need for further investigation in this area. 

RFA \cite{awais2021adversarial} leverages a pretrained ImageNet model to develop a regularization term. 
When the trainable model is initialized with a pretrained robust model, the regularization term encourages the outputs of each layer of the trainable model to resemble those of the pretrained model.
This facilitates the learning of robust features. 
This approach can be applied to existing UDA algorithms such as DANN \cite{ganin2016domain} and MDD \cite{zhang2019bridging}
to make them adversarially robust.

ARTUDA \cite{lo2022exploring} adopts a self-supervised adversarial training approach, where the predictions
of the current model are used as pseudo-labels to generate adversarial examples for the target domain. 
It also generates adversarial examples for the source domain and applies the distance metrics used in DANN. 
However, there is no theoretical justification of the complex loss function
used in ARTUDA.
One key challenge in ARTUDA is that pseudo-labels are generated by the same model being trained. 
As a result, due to the inherent trade-off between robustness and generalization, as the model learns more robust features, pseudo-labels are getting more inaccurate. 
This issue becomes particularly pronounced when the perturbation budget ($\varepsilon$) is large.

SRoUDA \cite{zhu2023srouda} attempts to resolve this limitation by introducing a separate teacher model to assign pseudo-labels to the target data. 
By leveraging meta-learning, SRoUDA improves the accuracy of pseudo-labeling by the teacher model and dynamically performs adversarial training \cite{madry2018towards} on the target domain using these pseudo-labels. 
Despite the improvement, SRoUDA struggles to effectively capture domain-invariant features, as it does not explicitly consider the deviation of the distributions  between the source and target domains. 
This leads to a decline in performance in the source as well as in the unseen domains, indicating a failure to capture what we call essentially domain-invariant robust features. 
\section{Proposed Method}
\label{sec:proposed_method}
In this section, we present the theoretical results for adversarially robust UDA. 
This is an extension of MDD \cite{zhang2019bridging} to its robust version, and the proofs are deferred to \cref{sec:theo_res}.
\subsection{Robust Margin Disparity Discrepancy}
In this subsection, 
we introduce a novel disparity measure and key propositions to establish the generalization bound.

Denote the classifier $h_f$ for a given score function $f$ as  
\begin{equation*}
h_f: \bm{x} \mapsto \underset{y \in \mathcal{Y}}{\argmax} f_y(\bm{x}),    
\end{equation*}
and denote the margin of $f$ at a labeled sample $(\bm{x}, y)$ as  
\begin{equation*}
\mathcal{M}_f(\bm{x}, y) = f_y(\bm{x}) - \underset{y' \ne y}{\max} f_{y'}(\bm{x}).
\end{equation*}
\begin{definition}(\textbf{Robust Disparity}).
\label{rob_disparity}
For given $f ,{f'} \in \mathcal{F}$, the \textit{0-1 robust disparity} between $f$ and $f'$ on a distribution $\mathcal{D}_\mathbf{X}$ on $\mathcal{X}$ is defined by
\begin{equation}
\operatorname{disp}^{\text{rob}}_{\mathcal{D}_\mathbf{X}}(f', {f}) := \mathbb{E}_{\mathbf{X} \sim \mathcal{D}_\mathbf{X}}
\underset{\mathbf{X}' \in \mathcal{B}_p (\mathbf{X}, \varepsilon)}{\max} \mathbbm{1}\{ h_{f'}(\mathbf{X}) \neq h_{f}(\mathbf{X}')\}
\end{equation}
\end{definition}
The robust disparity is measured based on the adversarial robustness of the second model $f$,
while the first model $f'$ serves as a pseudo-label generator.
Note that if the prediction of $f'$ is consistent to true label, it is the robust risk as defined in \cref{eq:rob_risk}.

As an extension of the margin loss in the non-robust setting \cite{zhang2019bridging} to its robust counterpart, we define the \textit{robust margin loss} as
\begin{equation}
\mathcal{R}^{\text{rob}, (\rho)}_{\mathcal{D}}(f):=\mathbb{E}_{(\mathbf{X},Y)\sim\mathcal{D}}\underset{\mathbf{X}'\in\mathcal{B}_p(\mathbf{X},\varepsilon)}{\max} \Phi_{\rho} \circ \cm_f(\mathbf{X}',Y),
\end{equation}
where 
\begin{equation}
\Phi_\rho(x) := \begin{cases}0 & \rho \leq x \\ 1-x / \rho & 0 \leq x \leq \rho \\ 1 & x \leq 0
\end{cases}
\end{equation}
and $\circ$ is composite operator of functions.

\begin{definition}(\textbf{Robust Margin Disparity}). 
\label{rob_margin_disparity}
    For given $f,f'\in\mathcal{F}$, the Robust Margin Disparity is defined as
    \begin{equation}
    \begin{aligned}
        &\operatorname{disp}^{\text{rob}, (\rho)}_{\mathcal{D}_\mathbf{X}}(f', f)\\
        &:=\mathbb{E}_{\mathbf{X} \sim \mathcal{D}_\mathbf{X}} \underset{\mathbf{X}' \in \mathcal{B}_p(\mathbf{X}, \varepsilon)}{\max} \Phi_{\rho} \circ \cm_f(\mathbf{X}', h_{f'}(\mathbf{X}))
    \end{aligned}
    \end{equation}
\end{definition}
An important property of the robust margin disparity is that it is greater than the robust disparity, i.e. 
$\operatorname{disp}_{\mathcal{D}_\mathbf{X}}^{\text{rob},(\rho)}(f',f)\ge\operatorname{disp}_{\mathcal{D}_\mathbf{X}}^{\text{rob}}(f',f)$. 

We now use this disparity term to bound the robust risk on the target distribution with additional feasible terms.
\begin{restatable}{proposition}{tarotprop}
\label{prop:upper_bound}
    Let $\mathcal{S}$ and $\mathcal{T}$ represent the distributions of the source and target domains, respectively. Similarly, let $\mathcal{S}_\mathbf{X}$ and $\mathcal{T}_\mathbf{X}$ denote the marginal distributions of the source and target domains over $\mathbf{X}$, respectively.
    For every score function $f\in\mathcal{F}$, the following inequality holds: 
    \begin{equation}
    \begin{aligned}
        &\mathcal{R}_\mathcal{T}^{\text{rob}}(f)\le\\
        &\mathcal{R}_\mathcal{S}^{(\rho)}(f)+\left\{ \operatorname{disp}^{\text{rob},(\rho)}_{\mathcal{T}_\mathbf{X}}(f^*,f)-\operatorname{disp}^{(\rho)}_{\mathcal{S}_\mathbf{X}}(f^*,f) \right\}+\lambda,
    \end{aligned}
    \end{equation}
    where $f^*=\underset{f\in\mathcal{F}}{\argmin}\{\mathcal{R}_\mathcal{T}^{(\rho)}(f)+\mathcal{R}_\mathcal{S}^{(\rho)}(f)\}$ is ideal hypothesis 
    and 
    $\lambda=\lambda(\mathcal{F},\mathcal{S},\mathcal{T},\varepsilon,\rho)=\mathcal{R}_\mathcal{T}^{(\rho)}(f^*)+\mathcal{R}_\mathcal{S}^{(\rho)}(f^*)$ is constant of $f$.
\end{restatable}
Note that the upper bound consists the standard margin risk of the source domain, not the robust margin risk. 
This enables the corresponding algorithm introduced in \cref{sec:learning_algorithm} computationally simpler
because generation of adversarial samples is not required to calculate the standard margin risk on the source domain. 

From the above proposition, we can naturally induce the following discrepancy between the distributions of two domains, which we call as a robust margin disparity discrepancy. 
\begin{definition}(\textbf{Robust Margin Disparity Discrepancy}). 
\label{def:rob_margin_dd}
Given a score function $f\in\mathcal{F}$, the Robust Margin Disparity Discrepancy induced by $f' \in \mathcal{F}$ of two distributions $\mathcal{S}_\mathbf{X}$ and $\mathcal{T}_\mathbf{X}$ is defined as follows: 
\begin{equation}
\begin{aligned}
    &\mathit{d}^{\text{rob}, (\rho)}_{f,\mathcal{F}}(\mathcal{S}_\mathbf{X},\mathcal{T}_\mathbf{X})\\
    &:=\underset{f'\in\mathcal{F}}{\sup}\left\{ \operatorname{disp}^{\text{rob}, (\rho)}_{\mathcal{T}_\mathbf{X}}(f',f) - \operatorname{disp}^{ (\rho)}_{\mathcal{S}_\mathbf{X}}(f',f) \right\}
\end{aligned}
\end{equation}
\end{definition}

This discrepancy measures the difference of the robust margin disparity on
target domain and the standard margin disparity on the source domain,
where the objective $f$ is fixed. If there is significant gap between the adversarial target domain and standard source domain, the discrepancy becomes  large.
Now we introduce a non-asymptotic upper bound for the target robust risk:
\begin{equation}
\label{eq:upper_bound}
    \mathcal{R}_\mathcal{T}^{\text{rob},(\rho)}(f)\le\mathcal{R}_\mathcal{S}^{(\rho)}(f)+d_{f,\mathcal{F}}^{\text{rob},(\rho)}(\mathcal{S}_\mathbf{X},\mathcal{T}_\mathbf{X})+\lambda
\end{equation}
with the same $\lambda$ in \cref{prop:upper_bound}.

\subsection{Generalization bound}
In this subsection, we provide a generalization bound based on \cref{eq:upper_bound}.
We use the notion of local Lipschitz constant of a function, following \citet{yang2020closer}.
\begin{definition}[\textbf{Local Lipschitz Constant}]
    Consider an arbitrary real function $f:\mathbb{R}^d\to\mathbb{R}^C$, and a $L_p$-norm $\lVert\cdot\rVert_p$. 
    Define a point-wise local Lipschitz constant as 
    \begin{equation}
        L_f(\bm{x}, \varepsilon):=\underset{\bm{x}' \in \mathcal{B}_p(\bm{x}, \varepsilon)}{\sup}\frac{\lVert f(\bm{x}')-f(\bm{x}) \rVert_1}{\lVert \bm{x}'-\bm{x} \rVert_p}
    \end{equation}
    for a point $\bm{x}$.
    Then, we define a local Lipschitz constant on $\mathcal{D}_\mathbf{X}$ as 
    \begin{equation}
        L_f(\mathcal{D}_\mathbf{X},\varepsilon):=\underset{x \in \mathcal{D}_\mathbf{X}}{\sup}L_f(\bm{x},\varepsilon)
    \end{equation}
    We call that $f$ is locally Lipschitz, if $L_f(\mathcal{D}_\mathbf{X},\varepsilon)$ is bounded (i.e. $L_f(\mathcal{D}_\mathbf{X},\varepsilon) < \infty$). 
\end{definition}
We also define following auxiliary function spaces. 
\begin{definition}
    Consider a set of score functions $\mathcal{F}$ and
    let $\mathcal{H}$ be the class of the induced classifiers $h_f$ for each $f\in\mathcal{F}$. 
    We define the following two sets. 
    \begin{align}
        \Pi_\mathcal{H}\mathcal{F}&:=\{ x\mapsto f(x,h(x))\vert h\in\mathcal{H},f\in\mathcal{F} \}\\
        \Pi_1\mathcal{F}&:=\{ x\mapsto f(x,y)\vert y\in\mathcal{Y},f\in\mathcal{F} \}
    \end{align}
\end{definition}
It is known that $\Pi_\mathcal{H}\mathcal{F}$ takes a form of the space of inner products of vector fields from $\mathcal{H}$ to $\mathcal{F}$ \cite{galbis2012vector,zhang2019bridging}. 
The notion $\Pi_1\mathcal{F}$ follows from \citet{mohri2012foundation}. 


We now derive a generalization bound for the robust risk on the target domain using Rademacher complexity, which quantifies the expressiveness of a given hypothesis space \cite{mohri2012foundation}.

\begin{definition}(\textbf{Rademacher Complexity}).
Let $\mathcal{F}$ be a familiy of functions mapping from $\mathcal{Z}=\mathcal{X} \times \mathcal{Y}$ to [a, b] and $\widehat{\mathcal{D}}=\{z_1, \cdots, z_n\}$ are finite samples independently drawn from a distribution $\mathcal{D}$ on $\mathcal{Z}$. 
Then, the empirical Rademacher complexity of $\mathcal{F}$ with $\widehat{\mathcal{D}}$ is defined by 
\begin{equation}
\widehat{\mathfrak{R}}_{\widehat{\mathcal{D}}}(\mathcal{F}) \triangleq \mathbb{E}_\sigma \sup _{f \in \mathcal{F}} \frac{1}{n} \sum_{i=1}^n \sigma_i f\left(z_i\right),
\end{equation}
where $\sigma_i$s are independent random variables following the Rademacher distribution.
In turn, the Rademacher complexity of $\mathcal{F}$ is defined by
\begin{equation}
\mathfrak{R}_{n,\mathcal{D}}(\mathcal{F}) \triangleq \mathbb{E}_{\widehat{\mathcal{D}} \sim \mathcal{D}^n} \widehat{\mathfrak{R}}_{\widehat{\mathcal{D}}}(\mathcal{F}) 
\end{equation}
\end{definition}

Now we provide our generalization bound, which replaces the terms
in the upper bound (\cref{eq:upper_bound}) that depend on the population distribution by
the terms depending on the empirical distributions.

\begin{restatable}{theorem}{tarotgb}(\textbf{Generalization Bound on the Robust Risk of Target Distribution}).
\label{tarot_gb}
For any $\delta > 0 $, with probability $1-3\delta$, we have the following uniform generalization bound for any score function $f$ in $\mathcal{F}:$
\begin{equation}
\begin{aligned}
    &\mathcal{R}^{\text{rob}}_{\mathcal{T}}(f)\\
    & \le \mathcal{R}^{(\rho)}_{\widehat{\mathcal{S}}}(f) + \text{d}^{\text{rob}, (\rho)}_{f, \mathcal{F}}(\widehat{\mathcal{S}}_\mathbf{X}, \widehat{\mathcal{T}}_\mathbf{X}) + \lambda\\
    &+ \frac{2C^2}{\rho}\mathfrak{R}_{n, \mathcal{S}}(\Pi_1 \mathcal{F}) + \frac{2C}{\rho}\mathfrak{R}_{n, \mathcal{S}}(\Pi_{\mathcal{H}} \mathcal{F}) + 2\sqrt{\dfrac{\log 2/\delta}{2n}}\\
    &+ \frac{2C}{\rho}\mathfrak{R}_{m, \mathcal{T}}(\Pi_\mathcal{H} \mathcal{F}) + \sqrt{
    \dfrac{\log 2/\delta}{2m}}
    +\frac{2\varepsilon L_f(\mathcal{T}_\mathbf{X},\varepsilon)}{\rho},
\end{aligned}
\end{equation}
where $\lambda=\underset{f\in\mathcal{F}}{\min}\{\mathcal{R}_\mathcal{T}^{(\rho)}(f)+\mathcal{R}_\mathcal{S}^{(\rho)}(f)\}$. 
\end{restatable}

\subsection{Two insights}
In this subsection, 
we present two key insights for robust UDA. First, we analyze the impact of the local Lipschitz constant on generalization performance, showing why initializing robust UDA algorithms with a robustly pre-trained (Robust-PT) model can substantially improve outcomes. 
Second, we highlight why pseudo labeling alone may be insufficient, especially in large or complex target domains where confirmation bias and the lack of explicit domain alignment can lead to persistent misclassifications.

\paragraph{About Local Lipschitz constant}
In \cref{tarot_gb}, there is the local Lipschitz constant $L_f(\mathcal{T}_\mathbf{X},\varepsilon)$. 
As discovered in previous literature, the local Lipschitz constant is known to be deeply involved with a generalization performance of a model \cite{bartlett2017spectrally,jordan2020exactly,shi2022efficiently}. 
This fact leads to various researches that investigate a method to insure provable robustness of a model \cite{cisse2017parseval,wu2021wider,tholeti2022robust}. 
It is also known that adversarial training empirically reduces the local Lipschitz constant of a neural network \cite{yang2020closer,zhang2022rethinking}.

In the numerical studies, we have found that robust UDA approaches with a Robust-PT as an initial
improve the prediction performance much compared to those without a Robust-PT as an initial. 
This phenomenon can be partly explained by the local Lipschitz term. 
That is, model which is not initialized with Robust-PT could have a large value of the local Lipschitz constant and thus could have a too large robust risk value on the target domain. 
Hence, using an initial solution with a small local Lipschitz constant is promising and a Robust-PT would be such one. 

\paragraph{Pseudo labeling only may not be sufficient}
Pseudo labeling is a widely used technique in domain adaptation \cite{lo2022exploring,zhu2023srouda}, where the objective is to improve the performance of the model on the target domain by iteratively using the predictive labels of the currently trained  model (or its variation)
as labels for the target domain data when generating adversarial examples for the target data.
There are at least two problems in those pseudo labeling approaches.

The first problem is a risk of confirmation bias. 
Since the model generates pseudo labels based on its own predictions, early-stage errors in the model could become self-reinforcing. 
When incorrect pseudo labels are assigned with high confidence, these mistakes persist through training cycles, amplifying errors and leading to poor generalization on the target domain. 
This phenomenon is widely studied in semi-supervised settings \cite{arazo2020pseudo,chen2023two}. 
Similarly in the UDA setting, as the training phase progresses, the model may increasingly rely on those incorrect predictions, making it difficult to recover especially in complex or noisy target domains. 

The risk of confirmation bias can be reduced by generating pseudo labels
from a pretrained non-robust UDA model called a (fixed) teacher model. 
In this case, we can write the objective function as
 \begin{equation}
    \mathbb{E}_{\mathbf{X}\sim\mathcal{T}_\mathbf{X}}\advmax\mathbbm{1}\left\{ h_f(\mathbf{X}')\ne h_{f^*}(\mathbf{X}) \right\}
\end{equation}
where $f^* = \underset{f\in\mathcal{F}}{\argmin} \mathcal{R}_\mathcal{\widehat{T}}(f)$.
In our notation, this term exactly coincides with $\operatorname{disp}_{\mathcal{T}_\mathbf{X}}^{\text{rob}}(f^*,f)$. 
The following proposition justifies this approach.

\begin{restatable}{proposition}{plub}
\label{prop:pseudo-label_ub}
    For every score function $f\in\mathcal{F}$, the following inequality holds: 
    \begin{equation}
        \mathcal{R}_\mathcal{T}^{\text{rob}}(f)\le\operatorname{disp}_{\mathcal{T}_\mathbf{X}}^\text{rob}(f^*,f)+\mathcal{R}_\mathcal{T}(f^*)
    \end{equation}
    for arbitrary $f^*\in\mathcal{F}$. 
\end{restatable}
It can be easily seen through the intermediate steps of \cref{prop:upper_bound}.
By letting $f^*$ as a non-robust UDA model, the pseudo labeling approach
with the teacher model $f^*$ can be understood to minimize the upper bound in \cref{prop:pseudo-label_ub}.
Our empirical studies amply confirm this insight. 

Pseudo labeling approaches with a fixed teacher model, however, still has limitations. 
In particular, the scalability of pseudo labeling across diverse or large datasets can be challenging due to the absence of explicit domain alignment mechanisms. As the size or complexity of the target domain grows, the chance of 
incorrect pseudo labels increases, which could overwhelm the overall learning process.
This may happen especially in scenarios where the source and target domains differ significantly. This makes it harder for the model to generalize to more extensive or varied datasets, particularly when the domain gap becomes larger due to increasing size or diversity of data.

Note that our generalization bound \cref{tarot_gb} uses the robust margin disparity discrepancy between
the source and target domains instead of the robust disparity only on the target domain in 
\cref{prop:pseudo-label_ub}. That is, our generalization bound integrates pseudo labeling and domain alignment
together. 
In the following subsection, inspired by these insights, we introduce an algorithm designed to overcome the drawbacks of pseudo labeling approaches, tackling both confirmation bias and scalability while enhancing generalization abilities across diverse and complex target domains.

\begin{algorithm}[h]
    \small
    \caption{Transferring Adversarially Robust Training (TAROT)}
    \label{alg:tarot}
    \SetKwInOut{Input}{Input}
    \SetKwInOut{Output}{Output}
    \SetKw{Return}{Return}
    \Input{$\psi$ : feature extractor, $\pi$ : head,  $\pi'$ : auxiliary head, 
    $s$ : standard trained model, $\mathcal{D}_{\mathcal{S}}=\{(\bm{x}^{\text{s}}_i, y^\text{s}_i)\}_{i=1}^{m}:$
    source domain dataset,
     $\mathcal{D}_{\mathcal{S}}=\{(\bm{x}^{\text{s}}_i, y^\text{s}_i)\}_{i=1}^{m}:$$\mathcal{D}_{\mathcal{T}}=
     \{\bm{x}^\text{t}_j \}_{j=1}^{m'}:$ target domain dataset, 
    $\eta_1, \eta_2$, $\lambda$, $\rho:$ learning rates in \cref{objective-tarot}, $T:$ the number of epochs,
    $B:$ the number of batches, $K:$ batch size}
    \Output{adversarially robust network $\pi \circ \psi$}
    $\bm{\theta}_{\psi} \leftarrow \bm{\theta}_{\text{R}}$ (initialized by a robustly trained feature extractor)
    
    \For{$ b = 1 , \cdots, B$}{
        \For{$ k = 1 , \cdots, K$}{
        Generate $\widehat{\bm{x}}^{\text{t}, \text{adv}}_{b, k}$ by $\text{PGD}(\pi \circ \psi  (\bm{x}^{t}_{b, k}), \tilde{y}^{t}_{b,k}),$ $\; \tilde{y}^{t}_{b,k} = h_{s} (\bm{x}^{t}_{b,k}) $\\
        Update 
        \footnotesize
        $(\bm{\theta}_{\psi}, \bm{\theta}_{\pi}) \leftarrow (\bm{\theta}_{\psi}, \bm{\theta}_{\pi}) -\eta_1\frac{1}{K} \nabla_{(\bm{\theta}_{\psi}, \bm{\theta}_{\pi})}\mathcal{\ell}_{\text{TAROT}}({\bm{\theta}} ; \{(\bm{x}^{\text{s}}_{b, k}, y^{\text{s}}_{b, k})\}_{k=1}^{K}, \{(\bm{x}^{\text{t}}_{b, k})\}_{k=1}^{K})$\\
        \normalsize
        Update 
        \footnotesize
        $\bm{\theta}_{\pi'} \leftarrow \bm{\theta}_{\pi'} + \eta_2\frac{1}{K} \nabla_{\bm{\theta}_{\pi'}}\mathcal{\ell}_{\text{TAROT}}({\bm{\theta}} ; \{(\bm{x}^{\text{s}}_{b, k}, y^{\text{s}}_{b, k})\}_{k=1}^{K}, \{(\bm{x}^{\text{t}}_{b, k})\}_{k=1}^{K} , \lambda)$
        }
    }
    \Return{$\pi \circ \psi$}
\end{algorithm}

\subsection{Learning Algorithm}
\label{sec:learning_algorithm}
Inspired by \cref{tarot_gb}, we propose the robust domain adaption algorithm, which we call
\textbf{T}ransferring \textbf{A}dversarially \textbf{RO}bust \textbf{T}raining (TAROT).
Our basic idea is to find a optimal score function $f\in\mathcal{F}$ by solving
a modified version of the following optimization problem:
\begin{equation}
\label{optim}
    \underset{f\in\mathcal{F}}{\min}\;\mathcal{R}_{\widehat{\mathcal{S}}}^{(\rho)}(f)+d_{f,\mathcal{F}}^{\text{rob},(\rho)}(\widehat{\mathcal{S}}_\mathbf{X},\widehat{\mathcal{T}}_\mathbf{X}) + L_f(\widehat{\mathcal{T}}_\mathbf{X},\varepsilon)
\end{equation}

\begin{itemize}
  \item As is done in DANN (Domain-Adversarial Neural Network) \cite{ganin2016domain} and MDD (Margin Disparity Discrepancy) \cite{zhang2019bridging}, 
  we consider a neural network structure that is composed of a feature extractor $\psi$ and a head $\pi$, i.e. a score function $f\in\mathcal{F}$ can be represented as $\pi\circ\psi$. 
  We only consider $\pi$ with $\psi$ being fixed, when calculating $d_{f,\mathcal{F}}^{\text{rob},(\rho)}$. 
  That is, we calculate the robust margin disparity discrepancy
  on the feature space of $\psi.$

    \item To reduce the first and second term, we follow the formulation of the MDD algorithm. 
    Likewise, we utilize an auxiliary head $\pi'$ and $\gamma:=\exp\rho$ designed to attain the margin. 
    $\pi'$ shares the same hypothesis space with $\pi$, and works as an adversarial network. 
    The discrepancy term, which is non-differentiable, can be optimized through GRL \cite{ganin2015unsupervised}. 
    The only difference is that, when to calculate $d_{f,\mathcal{F}}^{\text{rob},(\rho)}$, it requires calculating 
    a robust disparity and a standard disparity rather than two standard disparities. 
    \item To reduce the local Lipschitz constant, we minimize the cross entropy loss with adversarial examples for given pseudo labels attached by a teacher model learned by a nonrobust UDA approach. 
    This is based on the previously mentioned phenomenon that adversarial training reduces the local Lipschitz constant \cite{yang2020closer,zhang2022rethinking}. 
    \item When calculating the robust margin disparity discrepancy, instead of generating adversarial examples, we
    reuse those generated for the local Lipshitz constant, making the algorithm computationally effective. 
    \item As an initial model for $\psi$, we used a robustly pretrained models with from the repository provided in \cite{salman2020do}.
\end{itemize}

\begin{figure}[!t]
    \centering
    \includegraphics[width=1.\linewidth]{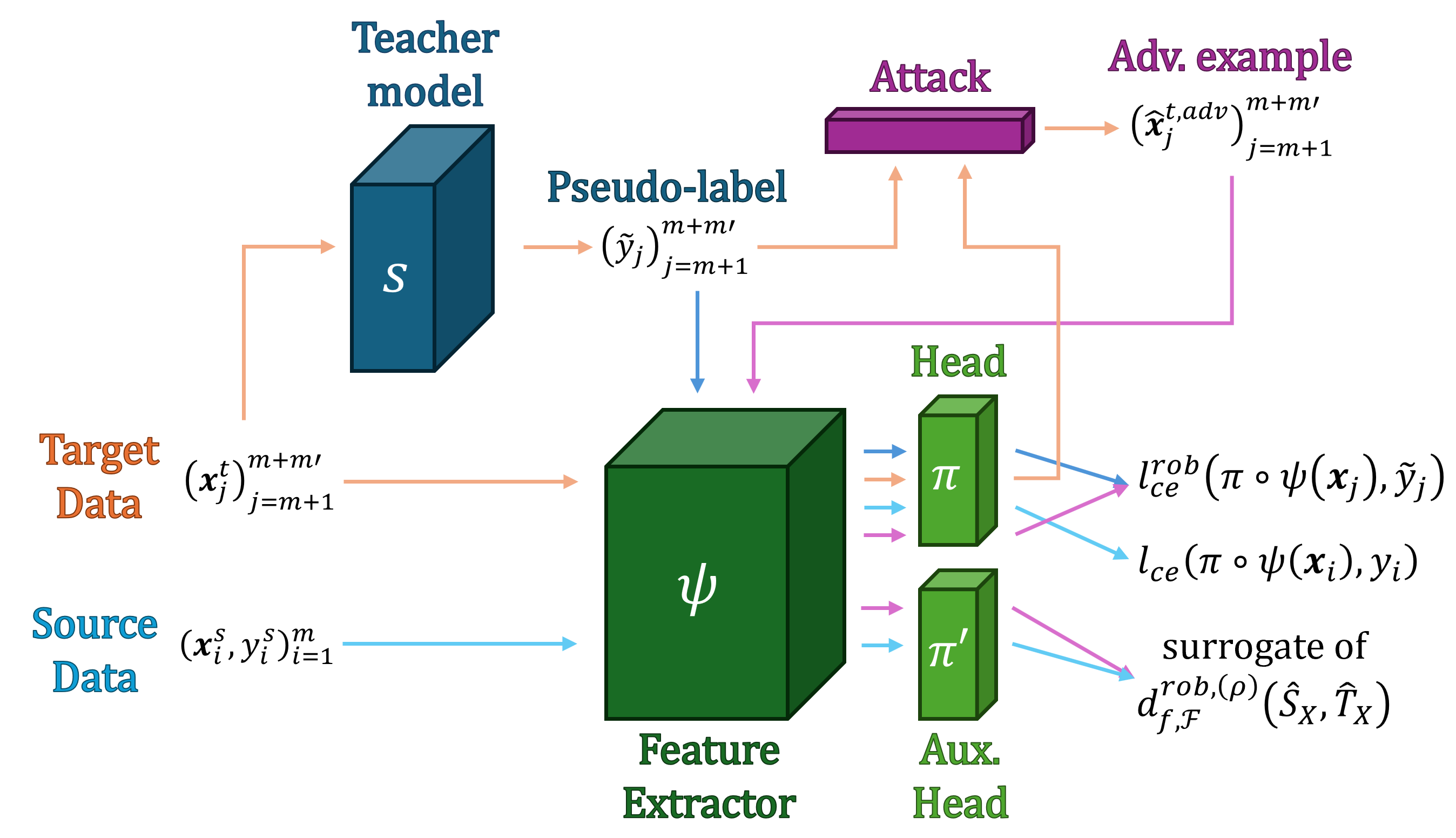}
    \caption{Overview of TAROT algorithm.}
    \label{fig:tarot}
\end{figure}

The objective function of TAROT is given as:
\begin{equation}
\begin{aligned}
\label{objective-tarot}
    &\mathcal{\ell}_{\text{TAROT}}({\bm{\theta}} ; \{(\bm{x}^{\text{s}}_{i}, y^{\text{s}}_{i})\}_{i=1}^{m}, \{(\bm{x}^{\text{t}}_{j})\}_{j=m+1}^{m+m'}, \alpha, \gamma)  \\ 
    & = \alpha \bigg\{\dfrac{1}{m} \sum\limits_{i=1}^m \ell_{\text{ce}} \left( ( \pi \circ \psi)(\bm{x}_i), y_i) \right) \\
    &\quad + \bigg( \dfrac{1}{m'} \sum\limits_{j=m+1}^{m+m'} \ell^{\text{rob}}_{\text{mod-ce}}((\pi' \circ \psi  )(\bm{x}_j),  h_{\pi \circ \psi}(\bm{x}_j)) \\
    &\quad- \dfrac{\gamma}{m} \sum\limits_{j=1}^{m} \ell_{\text{ce}}((\pi' \circ \psi )(\bm{x}_j), h_{\pi \circ \psi}(\bm{x}_j)) \bigg)\bigg\} \\
    &\quad+ \dfrac{1}{m'} \sum\limits_{j=m+1}^{m+m'} \ell^{\text{rob}}_{\text{ce}}((\pi \circ \psi  )(\bm{x}_j),  \tilde{y}_j)
\end{aligned}
\end{equation}
where $\ell_\text{ce}$, $\ell_\text{mod-ce}^\text{rob}$ \cite{zhang2019bridging} and
$\ell_\text{ce}^\text{rob}$ \cite{madry2018towards} 
as cross-entropy, modified cross-entropy with adversarial examples and cross-entropy with adversarial examples
respectively (See \cref{app:details}). 
Also,
$\bm{\theta}=(\bm{\theta}_\psi,\bm{\theta}_\pi,\bm{\theta}_{\pi'})$ to represent the parameters corresponding to $\psi,\pi,\pi'$.
$\tilde{y}_j=h_s(\bm{x}_j)$ is a pseudo label generated by a standard trained model $s$ and
$\alpha$ is a trade-off parameter that balances transferability and target's robustness.
TAROT algorithm is summarized in \cref{alg:tarot} and \Cref{fig:tarot}. 

\section{Experiments}
\label{sec:experiments}
In this section, we compare TAROT with other competitors for adversarially robust UDA. 
We use several benchmark datasets: Office-31 \cite{saenko2010adapting}, Office-Home \cite{venkateswara2017deep}, VisDA2017 \cite{visda2017}, and DomainNet \cite{peng2019moment}. 
Throughout these datasets, effectiveness of TAROT is evaluated across a wide range of domain shifts and robustness scenarios.
Further experimental details including explanation for datasets and experimental setups are presented in \cref{app:details}. 
The code is available in the repository$^{\ddagger}$ \def\thefootnote{$\ddagger$}\footnotetext{\url{https://github.com/dyoony/TAROT}}.

\begin{table*}[!t]
    \caption{\textbf{Performances of ARTUDA, RFA, SRoUDA, PL and TAROT on Office31 ($\varepsilon=16/255$).} In each cell, the first number is the standard accuracy (\%), while the second number is the robust accuracy (\%) for AA. Bold numbers indicate the best performance.
    }
    \footnotesize
    \centering
    \begin{tabular}{c|cccccc|c}
    \specialrule{1pt}{0pt}{0pt}
    \textbf{Method} & A $\rightarrow$ D  & A $\rightarrow$ W & D $\rightarrow$ A & D $\rightarrow$ W & W $\rightarrow$ A & W $\rightarrow$ D & Avg.\\
    \specialrule{1pt}{0pt}{0pt}
    ARTUDA & 23.49 / 11.65 & 22.39 / 9.56 	& 35.00 / 16.51 & 68.55 / 35.85 &	36.81 / 19.74 &	74.10 / 51.61 & 43.39 / 24.15 \\
    RFA & 78.72 / 1.21 & 71.20 / 1.26 & 55.84 / 18.03 & 96.73 / 6.42 & 60.10 / 23.00 & 99.40 / 6.02 & 77.00 / 9.32 \\
    SRoUDA & 93.37 / 42.37 	& 78.87 / 65.16 & 17.93 / 3.44 & 96.00 / 44.28 & 66.53 / 47.21 & \textbf{100.00} / 62.45 	& 74.62 / 44.15 \\
    PL & 91.77 / \textbf{84.14} & 91.82 / 91.70 	& 74.19 / \textbf{67.87} & \textbf{97.99} / 95.98 & 72.42 / 65.46 & \textbf{100.00} / \textbf{92.57} & 88.03 / 82.95 \\
    \hline
    TAROT & \textbf{94.18} / \textbf{84.14} & \textbf{95.47} / \textbf{93.46} & \textbf{74.51} / 67.13 & \textbf{97.99} /	\textbf{96.23} & \textbf{73.73} / \textbf{66.88}  & \textbf{100.00} / 91.77  & \textbf{89.31} / \textbf{83.27}  \\
    \specialrule{1pt}{0pt}{0pt}
    \end{tabular}
    \label{table:office31}
    \vskip -0.05in
\end{table*}
\begin{table*}[!t]
    \caption{\textbf{Performances of ARTUDA, RFA, SRoUDA, PL and TAROT on OfficeHome ($\varepsilon=16/255$).} In each cell, the first number is the standard accuracy (\%), while the second number is the robust accuracy (\%) for AA. Bold numbers indicate the best performance.
    }
    \footnotesize
    \centering
    \begin{tabular}{c|cccccc|c}
    \specialrule{1pt}{0pt}{0pt}
    \textbf{Method} & Ar $\rightarrow$ Cl  & Ar $\rightarrow$ Pr & Ar $\rightarrow$ Rw & Cl $\rightarrow$ Ar & Cl $\rightarrow$ Pr & Cl $\rightarrow$ Rw & \\
    \specialrule{1pt}{0pt}{0pt}
    ARTUDA & 30.29 / 13.56 & 26.18 / 4.75 & 5.88 / 0.00 & 3.42 / 0.00 & 25.34 / 10.48 & 20.98 / 5.53 \\
    RFA & 48.39 / 17.89 & 54.67 / 8.76 & 63.74 / 6.34 & 43.22 / 3.17 & 58.03 / 9.15 & 57.24 / 5.69 \\
    SRoUDA & 51.84 / 37.89 & \textbf{75.60} / 45.46 & 74.25 / 40.28 &  57.97 / 33.42 & 67.76 / 54.77 & 67.32 / 35.83 \\
    PL & \textbf{55.72} / 44.31 & 73.53 / 58.49 & 73.10 / 40.26 & 57.68 / 31.15 & 70.31 /	56.21 & 68.14 / 37.37 
 \\
    \hline
    TAROT & \textbf{55.72} / \textbf{46.53} & 73.85 / \textbf{58.93} & \textbf{77.78} / \textbf{42.62} & \textbf{60.12} / \textbf{34.69} & \textbf{73.08} / \textbf{58.14} & \textbf{69.93} / \textbf{38.95} \\
    \specialrule{1pt}{0pt}{0pt}
    & Pr $\rightarrow$ Ar  & Pr $\rightarrow$ Cl & Pr $\rightarrow$ Rw & Rw $\rightarrow$ Ar & Rw $\rightarrow$ Cl & Rw $\rightarrow$ Pr & Avg.\\
    \specialrule{1pt}{0pt}{0pt}
    ARTUDA  & 8.74 / 1.73 & 38.92 / 14.57 & 47.53 / 5.30 & 30.12 / 2.10 & 41.79 / 22.29 & 45.19 / 13.97 & 27.03 / 7.86 \\
    RFA & 42.03 / 2.56 	& 47.86 / 13.31 	& 63.44 / 5.21 & 54.27 / 3.38 &	54.27 / 16.70 &	72.85 / 9.78 & 55.00 / 8.49 \\
    SRoUDA & 59.21 / 31.40 & 48.52 / 39.77  & 75.40 / 41.20 & 70.87 / 40.38 & 53.40 / 42.43 & 83.26 / 67.38 & 57.97 / 33.42 \\
    PL & 57.15 / 29.58 & 52.74 / 44.40 & 74.82 / 40.74 & 68.97 / 38.28 & 57.03 /	45.17 & 82.86 / 66.62 & 66.00 / 44.38 \\
    \hline
    TAROT & \textbf{59.62} / \textbf{32.76} & \textbf{54.98} / \textbf{47.01} & \textbf{78.66} / \textbf{42.94}  & \textbf{71.90} / \textbf{40.79} & \textbf{59.45} / \textbf{49.74} & \textbf{84.39} / \textbf{68.53} & \textbf{68.29} / \textbf{46.80} \\
    \specialrule{1pt}{0pt}{0pt}
    \end{tabular}
    \label{table:officehome}
    \vskip -0.15in
\end{table*}
\subsection{Performance Evaluation}
\paragraph{Training Setup} 

All of the results are averaged over three runs, with different teacher model initializations.
Different perturbation budgets ($\varepsilon$) are chosen based on the difficulty of each benchmark dataset. For more challenging datasets, a smaller 
$\varepsilon$ is used, and vice versa. Specifically, we use 
$\varepsilon=16/255$ for the Office-31 and Office-Home datasets, while 
$\varepsilon =8/255$ and 
$\varepsilon =4/255$ are used for VisDA2017 and DomainNet, respectively.
The values of 
$\alpha$ in \cref{objective-tarot} are set to 0.05, 0.1, 0.5, and 1.0 for Office31, OfficeHome, VisDA2017, and DomainNet,
respectively. 
We empirically observed that the optimal $\alpha$ values for each task vary much. 
The sensitivity analysis of $\alpha$ is provided in \cref{ablation}.

As baselines, along with ARTUDA \cite{lo2022exploring}, RFA \cite{awais2021adversarial}, and SRoUDA \cite{zhu2023srouda},
we consider the pseudo labeling (PL) algorithm which is
implemented as a simple PGD-AT algorithm 
where target pseudo labels are obtained from a fixed non-robustly trained teachers learned by MDD \cite{zhang2019bridging}. 
All algorithms are initialized with Robust-PT as is done for TAROT. 
Without Robust-PT, training with larger perturbations such as $\varepsilon=16/255$ can lead to non-decreasing training loss in certain tasks, as discussed in \cref{app:rob-PT}.

\vspace{-0.2in}
\paragraph{Evaluation Setup}

We report standard accuracies on clean samples and robust accuracies against AutoAttack \cite{croce2020reliable}. AutoAttack, which includes both white-box and black-box attacks, is a powerful evaluation tool that mitigates errors in robustness measurement caused by gradient obfuscation  —a phenomenon where model gradients are distorted or masked, leading to overestimated robustness \cite{athalye2018obfuscated}. AutoAttack is widely recognized for accurately assessing adversarial robustness by bypassing these gradient masking issues.


\subsubsection{Performance on Target Domains}
\label{subsec:performance_target}
\begin{figure*}[!ht]
    \centering
    \footnotesize
    \begin{minipage}{0.66\textwidth}\vspace{0pt}%
    \captionsetup{type=table}
    \caption{\textbf{Performances of ARTUDA, RFA, SRoUDA, PL and TAROT on DomainNet ($\varepsilon=4/255$).} 
    In each cell, the first number is the standard accuracy (\%), and the second number is the robust accuracy (\%) for AA. 
    Bold numbers indicate the best performance.
    }
    \begin{tabular}{@{}cc|cccc|c}
    \specialrule{.8pt}{0pt}{0pt}
    \multicolumn{2}{c|}{\textbf{Method}} & C $\rightarrow$ I  & C $\rightarrow$ P & C $\rightarrow$ R & C $\rightarrow$ S & Avg. \\
    \specialrule{.8pt}{0pt}{0pt}
    \multicolumn{2}{c|}{ARTUDA} & 12.64 / 3.76 & 31.15 / 7.90 & 44.90 / 16.49 & 39.90  / 17.58 & 32.15 / 11.43 \\
    \multicolumn{2}{c|}{RFA} & 9.42 / 3.47 	& 23.33 / 7.60 & 37.54 / 17.74 & 27.40 / 12.17 & 24.42 / 10.24 \\
    \multicolumn{2}{c|}{SRoUDA} & 9.27 / 5.68 & 24.33 / 15.07 & 42.91 / 28.38 & 32.42 / 23.82 & 27.23 /	18.24 \\
    \multicolumn{2}{c|}{PL} & 10.59 / 6.69 & 27.06 / 15.73 & 43.57 / 28.68 & 32.13 / 23.52 & 28.34 / 18.66 \\
    \hline
    \multicolumn{2}{c|}{TAROT} & \textbf{14.65} / \textbf{8.16} 	& \textbf{33.23} / \textbf{17.36} & \textbf{50.02} / \textbf{30.46} & \textbf{40.11} / \textbf{27.77} & \textbf{34.50} / \textbf{20.94} \\
    \specialrule{.8pt}{0pt}{0pt}
    \multicolumn{2}{c|}{} & I $\rightarrow$ C  & I $\rightarrow$ P & I $\rightarrow$ R & I $\rightarrow$ S &  Avg. \\
    \specialrule{.8pt}{0pt}{0pt}
    \multicolumn{2}{c|}{ARTUDA} & \textbf{31.89} / 12.07 & 20.38 / 2.56 & 23.79 / 5.47 & \textbf{34.61} / 5.90 & 27.67 / 6.50 \\
    \multicolumn{2}{c|}{RFA} & 15.00 / 8.82 & 	14.80 / 4.80 & 19.15 / 8.76 & 11.44 / 5.05 & 15.10 / 6.86 \\
    \multicolumn{2}{c|}{SRoUDA} & 25.95 / 14.18 & 24.27 / 13.93 & 36.32 / 22.06 & 20.08 / 13.52 & 26.65 / 15.92 \\
    \multicolumn{2}{c|}{PL} & 17.45 / 13.16 & 22.42 / 13.41 & 32.50 / 21.94 & 18.25 / 13.14 & 22.65 / 15.41 \\
    \hline
    \multicolumn{2}{c|}{TAROT} & 29.36 / \textbf{20.23} & \textbf{27.66} / \textbf{14.30} &	\textbf{39.14} / \textbf{23.39} &	21.89 / \textbf{14.08} & \textbf{29.51} / \textbf{18.00} \\
    \specialrule{.8pt}{0pt}{0pt}
    \multicolumn{2}{c|}{} & P $\rightarrow$ C  & P $\rightarrow$ I & P $\rightarrow$ R & P $\rightarrow$ S &  Avg. \\
    \specialrule{.8pt}{0pt}{0pt}
    \multicolumn{2}{c|}{ARTUDA} & \textbf{42.89} / 20.58 & 14.36 / 3.21 &	49.03 /	16.62 &	34.02 /	13.12 &	35.08 /	13.38 \\
    \multicolumn{2}{c|}{RFA} & 26.27 / 14.85 & 9.45 / 3.09 & 37.47 / 17.04 & 19.67 / 8.59 & 23.22 /	10.89  \\
    \multicolumn{2}{c|}{SRoUDA} & 36.69 / 28.26 & 10.17 / 5.86 & 45.69 / 29.95 & 30.95 / 19.14 & 30.87 / 20.80 \\
    \multicolumn{2}{c|}{PL} & 33.86 /	26.69 &	11.18 / 7.25 & 44.56 / 29.47 & 29.76 / 20.66 & 29.84 / 21.02 \\
    \hline
    \multicolumn{2}{c|}{TAROT} & 41.27 / \textbf{30.51} & \textbf{14.82} / \textbf{7.69} & \textbf{51.10} / \textbf{30.76} & \textbf{37.22} / \textbf{24.71} & \textbf{36.10} / \textbf{23.42} \\
    \specialrule{.8pt}{0pt}{0pt}
    \multicolumn{2}{c|}{} & R $\rightarrow$ C  & R $\rightarrow$ I  & R $\rightarrow$ P & R $\rightarrow$ S & Avg. \\
    \specialrule{.8pt}{0pt}{0pt}
    \multicolumn{2}{c|}{ARTUDA} & \textbf{52.94} / 28.10 & 16.94 / 4.17 &	41.95 /	12.02 &	36.11 / 14.86 &	36.98 /	14.79 \\
    \multicolumn{2}{c|}{RFA} & 36.72 / 19.30 & 12.36 / 3.02  & 32.03 / 8.34 & 22.73 / 8.12 & 25.96 	/ 9.70  \\
    \multicolumn{2}{c|}{SRoUDA} & 44.33 / 34.33 & 10.10 / 5.63 & 40.26 / 23.26 & 29.10 / 18.56 & 30.94 / 20.44 \\
    \multicolumn{2}{c|}{PL} & 42.03 / 33.27 & 11.85 / 7.14 & 39.26 / 22.80 & 29.05 / 19.68 & 30.55 / 20.72 \\
    \hline
    \multicolumn{2}{c|}{TAROT} & 52.26 / \textbf{37.58} & \textbf{18.86} / \textbf{9.45} & \textbf{47.62} / \textbf{23.43} & \textbf{39.38} / \textbf{25.23} & \textbf{39.53} / \textbf{23.92} \\
    \specialrule{.8pt}{0pt}{0pt}
    \multicolumn{2}{c|}{} & S $\rightarrow$ C  & S $\rightarrow$ I & S $\rightarrow$ P & S $\rightarrow$ R  & Avg. \\
    \specialrule{.8pt}{0pt}{0pt}
    \multicolumn{2}{c|}{ARTUDA} & \textbf{54.24} / 30.59 & 13.14 / 3.89 &	34.87 /	7.79 & 41.40 / 13.47 & 35.91 / 13.94 \\ \multicolumn{2}{c|}{RFA} & 33.87 / 19.94 & 	9.47 / 3.54 & 	27.29 / 7.83  & 30.39 / 13.70 & 25.26 / 11.25 \\
    \multicolumn{2}{c|}{SRoUDA} & 46.62 / 36.65 & 8.50 / 5.28 & 31.09 / 15.14 & 42.95 / 22.33 & 32.29 / 19.85\\
    \multicolumn{2}{c|}{PL} & 42.48 / 34.66  & 8.95 / 6.18 & 31.48 / 18.75 & 42.55 / 28.24 & 31.37 / 21.96 \\
    \hline
    \multicolumn{2}{c|}{TAROT} & 53.35 / \textbf{41.38} & \textbf{13.71} / \textbf{7.52} & \textbf{39.77} / \textbf{20.43} & \textbf{48.77} / \textbf{30.02} & \textbf{38.90} / \textbf{24.84} \\
    \specialrule{.8pt}{0pt}{0pt}%
    \label{table:domainnet}
    \end{tabular}
    \end{minipage}%
    \hspace{0.01\textwidth} 
    \begin{minipage}[!t]{0.3\textwidth}
    \vspace{-0.11in}%
    \footnotesize
    \captionsetup{type=table}
    \caption{\textbf{Performances of ARTUDA, RFA, SRoUDA, PL and TAROT on VisDA2017 ($\varepsilon=8/255$).} Standard accuracy (\%) / Robust accuracy (\%) for AA. 
    }
    \centering
    \begin{tabular}{@{}cc|cc@{}}
    \specialrule{.8pt}{0pt}{0pt}
    \multicolumn{2}{c|}{\textbf{Method}} & Syn. $\rightarrow$ Real & Syn. \\
    \specialrule{.8pt}{0pt}{0pt}
    \multicolumn{2}{c|}{ARTUDA} & 10.73 / 0.00 & 61.43 / 5.70 \\
    \multicolumn{2}{c|}{RFA} & 64.86 / 8.65 & \textbf{99.41} / 33.55  \\
    \multicolumn{2}{c|}{SRoUDA} & 24.44 / 0.20 & 48.56 / 24.05 \\
    \multicolumn{2}{c|}{PL} & 66.71 / \textbf{37.83} & 40.98 / 24.09 \\
    \hline
    \multicolumn{2}{c|}{TAROT} & \textbf{67.12} / 37.77 & 85.18 / \textbf{51.21} \\
    \specialrule{.8pt}{0pt}{0pt}
    \label{table:visda}
    \end{tabular}%
    \vspace{0.08in}
    \centering
    \includegraphics[width=5.6cm]{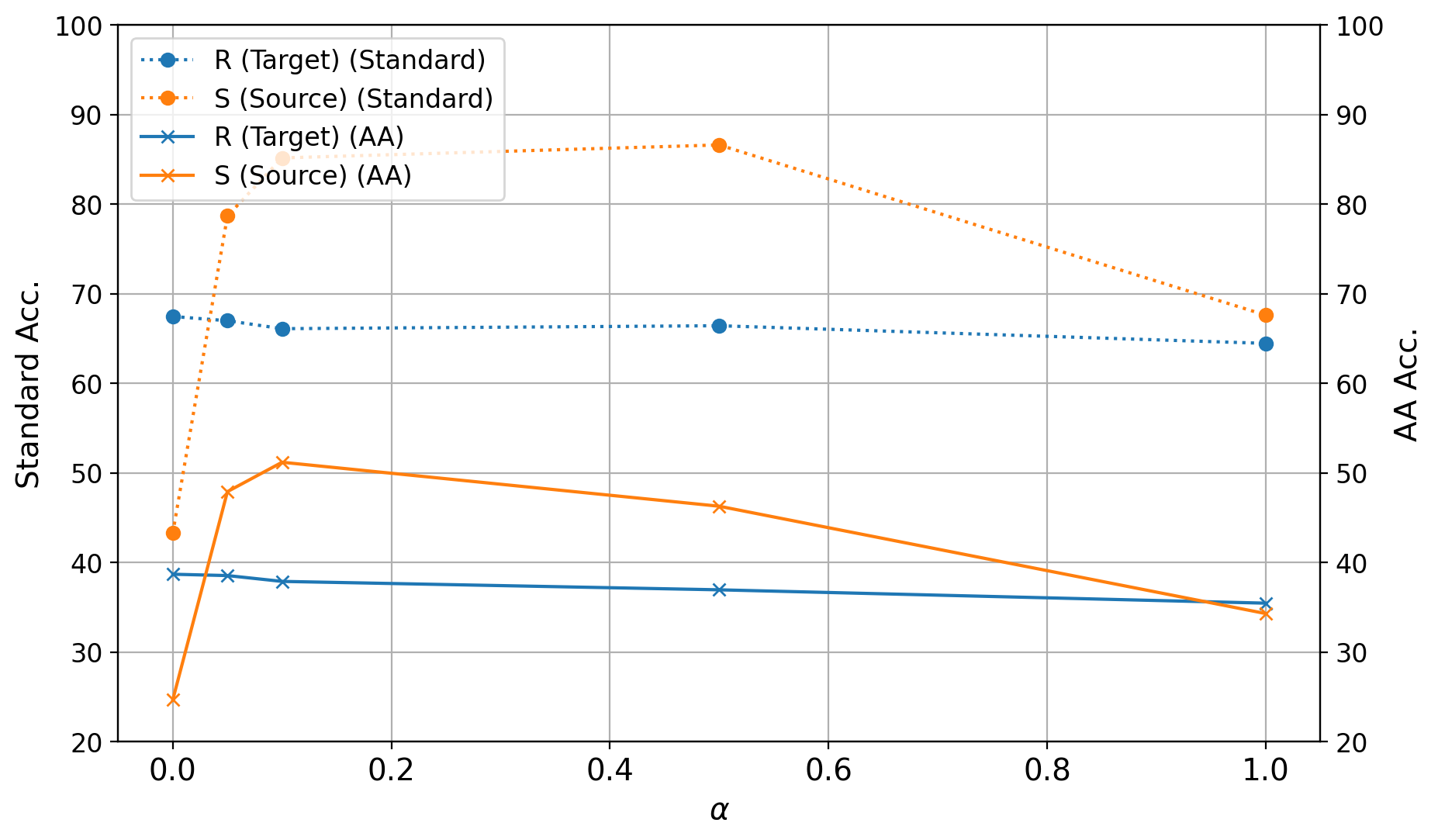}
    \includegraphics[width=5.6cm]{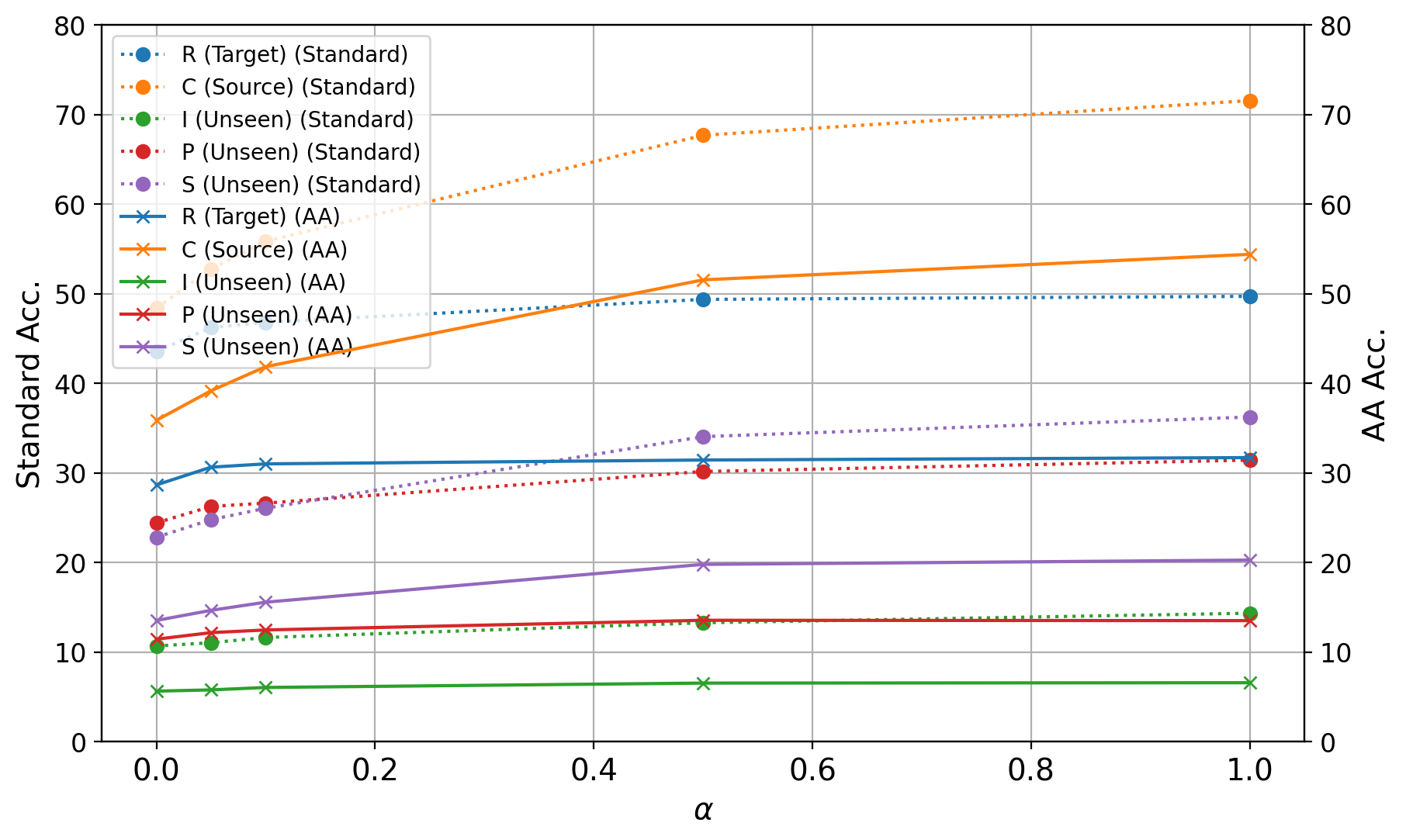}
    \vskip -0.1in
    \captionsetup{type=figure}
    \caption{Sensitivity Analysis of $\alpha$. $\alpha=0$ corresponds to PL.}
    \label{fig:sensitivity_alpha}
    \end{minipage}
    \vskip -0.2in
\end{figure*}

\Cref{table:office31,table:officehome,table:domainnet,table:visda} demonstrate that PL outperforms other algorithms, with the exception of TAROT.
As seen in \Cref{fig:robust_pt_effect}, simply combining Robust-PT as an initial solution with pseudo labeling can yield significant performance gains, which confirms our first theoretical insight. (See \cref{ablation}.)
This combination serves as a strong baseline in adversarially robust domain adaptation.
It's noteworthy that ARTUDA \cite{lo2022exploring}, RFA \cite{awais2021adversarial}, and SRoUDA \cite{zhu2023srouda} algorithms show remarkably poor performance on several tasks.
Note that the robust accuracies of a standard training algorithm (MDD) on the Office-31, Office-Home, and VisDA-2017 datasets are exactly 0.0, highlighting the importance of robust adaptation methods.
This is mainly attributed to our use of larger perturbation budgets during training and evaluation compared to those used in the original papers.
For reference, we provide results using smaller perturbation budgets ($4/255, 8/255$) on Office-31 and OfficeHome datasets in \cref{app:lower_perturb}.
Overall, TAROT outperforms existing algorithms across all benchmark datasets.
Note that the robust accuracies of a standard training algorithm (MDD) on the Office-31, Office-Home, and VisDA-2017 datasets are exactly 0.0.

\subsubsection{Essentially Domain-Invariant Robustness}

In this subsection, 
we empirically highlight the limitations of the pseudo labeling approach, which focuses solely on improving target domain performance without effectively learning domain-invariant features. We select tasks with Real world (Rw) as the target domain on the OfficeHome dataset. We compare TAROT (our algorithm), which is designed to reduce the distribution gap between source and target domains, with PL, evaluating both the source domain accuracy and the performance on unseen domains. \Cref{table:invariant-features} shows that our method not only adapts well to the target domain but also learns more generalizable and robust features, leading to improved robustness across domains by significant margins. 
We present an upper bound for robust risk on the source domain and demonstrate that the components of this bound closely align with those of the target domain, offering a partial theoretical explanation for TAROT's robustness to adversarial attacks on the source domain. 
Additional results for other tasks are provided in \cref{app:add_experiments}.
\begin{table}[!t]
    \caption{\textbf{Performances of PL and TAROT on Source Domain and Unseen Domain, on OfficeHome ($\varepsilon=8/255$).} In each cell, the first number is the standard accuracy (\%), while the second number corresponds to the robust accuracy (\%) for AA. Bold numbers indicate the best performance.
    A $\rightarrow$ B(C) indicates that A, B, and C represent the source, (unlabeled) target, and evaluation domains, respectively.}
    \scriptsize
    \centering
    \begin{tabular}{c|c|cc|c}
    \specialrule{.8pt}{0pt}{0pt}
     & \multicolumn{1}{c|}{\textbf{Source}} & \multicolumn{2}{c|}{\textbf{Unseen}}  \\
    \specialrule{.8pt}{0pt}{0pt}
    \textbf{Method} & Ar $\rightarrow$ Rw(Ar) & Ar $\rightarrow$ Rw(Cl) & Ar $\rightarrow$ Rw(Pr) & Avg. \\
    \hline
    ARTUDA & 18.83 / 2.64 & 7.70 / 0.89 & 6.28 / 0.45 & 10.94 / 1.33\\
    RFA & \textbf{99.62} / 47.02 & 42.52 / 21.53 & 51.57 / 21.51  & 64.57 / 30.02 \\ 
    SRoUDA & 39.39 / 17.43 & 39.54 / 28.66 & 50.75 / 35.86 & 42.23 / 27.32\\
    PL    & 41.78 / 19.41 &	41.97 /	31.39 &	52.42 /	37.24 &	45.39 /	29.34 \\
    TAROT & 98.35 /	\textbf{64.24} &	\textbf{47.86} /	\textbf{35.19} &	\textbf{58.32} /	\textbf{39.54} &	\textbf{68.18} /	\textbf{46.32} \\
    \specialrule{.8pt}{0pt}{0pt}
     & Cl $\rightarrow$ Rw(Cl) & Cl $\rightarrow$ Rw(Ar) & Cl $\rightarrow$ Rw(Pr) & Avg. \\
    \hline
    ARTUDA & 84.81 / 55.80 & 14.34 / 3.21 & 27.26 / 13.99 & 42.14 / 24.33 \\
    RFA & \textbf{98.35} / 80.10 & \textbf{40.37} / 8.86 & 54.86 / 22.25  & \textbf{64.53} / 37.07 \\ 
    SRoUDA & 48.29 / 35.58 & 33.87 / 15.62 & 48.19 / 33.07 &  43.45 / 28.09 \\
    PL    &  48.75 / 35.95 & 36.30 / 16.69 & 50.80 / 35.66 & 45.28 / 29.43 \\
    TAROT & 93.65 / \textbf{84.35} & 39.72 / \textbf{17.18} & \textbf{55.46} / \textbf{37.67} & 62.95 / \textbf{46.40} \\
    \specialrule{.8pt}{0pt}{0pt}
    & Pr $\rightarrow$ Rw(Pr) & Pr $\rightarrow$ Rw(Ar) & Pr $\rightarrow$ Rw(Cl) & Avg. \\
    \hline
    ARTUDA & 98.29 / 60.17 & 25.67 / 4.62 & 36.24 / 16.24 & 53.40 / 27.01\\
    RFA & \textbf{99.75} / 68.59 & 40.67 / 7.87 & 41.76 / 19.73 &  \textbf{60.73} / 32.06 \\ 
    SRoUDA & 61.48 / 41.40 & 34.82 / 16.07 & 42.08 / 30.68 & 46.13 / 29.38 \\
    PL    & 61.05 / 43.34 &	36.09 / \textbf{17.10} &	42.11 / 30.91 &	46.42 / 30.45 \\
    TAROT & 96.13 /	\textbf{82.43} &	\textbf{39.14} / 16.15 &	\textbf{46.87} / \textbf{33.01} &	60.71 / \textbf{43.86} \\
    \specialrule{.8pt}{0pt}{0pt}
    \end{tabular}
    \label{table:invariant-features}
    \vskip -0.2in
\end{table}

\subsection{Ablation Studies}
\label{ablation}
We conduct ablation studies about: (i) effect of Robust-PT (ii) sensitivity analysis of $\alpha$. 
All exact values and additional experimental results are presented in \cref{app:add_experiments}.

\begin{figure}[!ht]
    \centering
    \small
    \begin{subfigure}[t]{0.45\columnwidth}
        \centering
        \includegraphics[width=\textwidth]{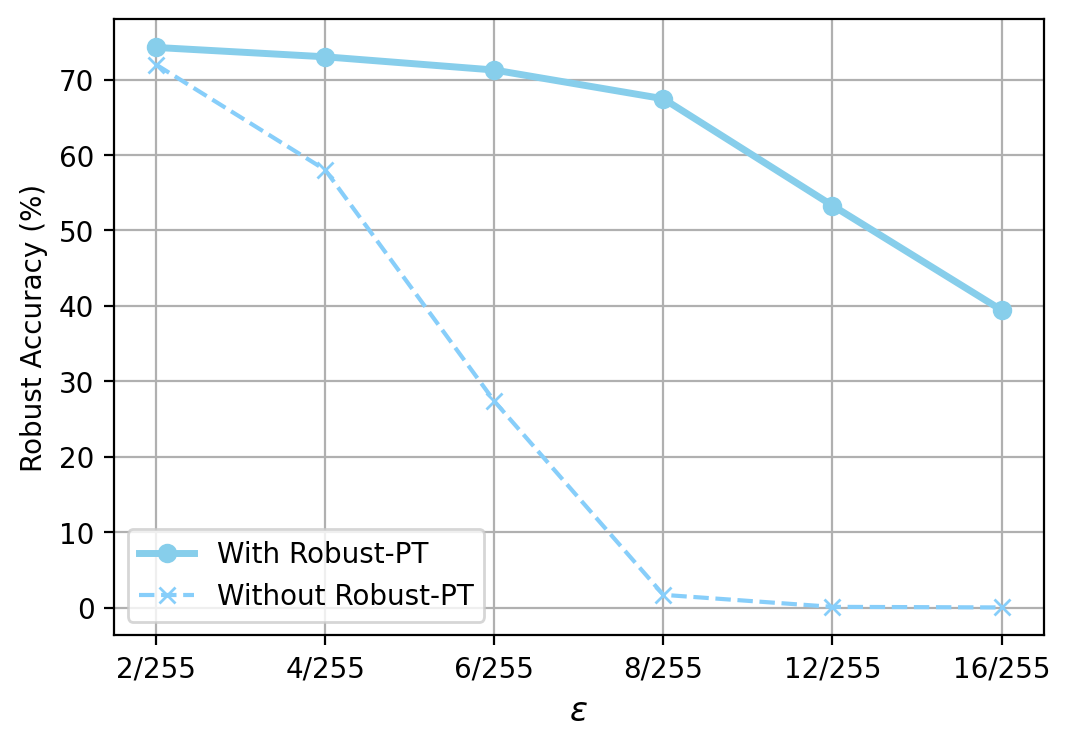}
        \caption{PL w/ and w/o R-PT}
        \label{fig:robust_pl_mdd}
    \end{subfigure} 
    \hfill
    \begin{subfigure}[t]{0.45\columnwidth}
        \centering
        \includegraphics[width=\textwidth]{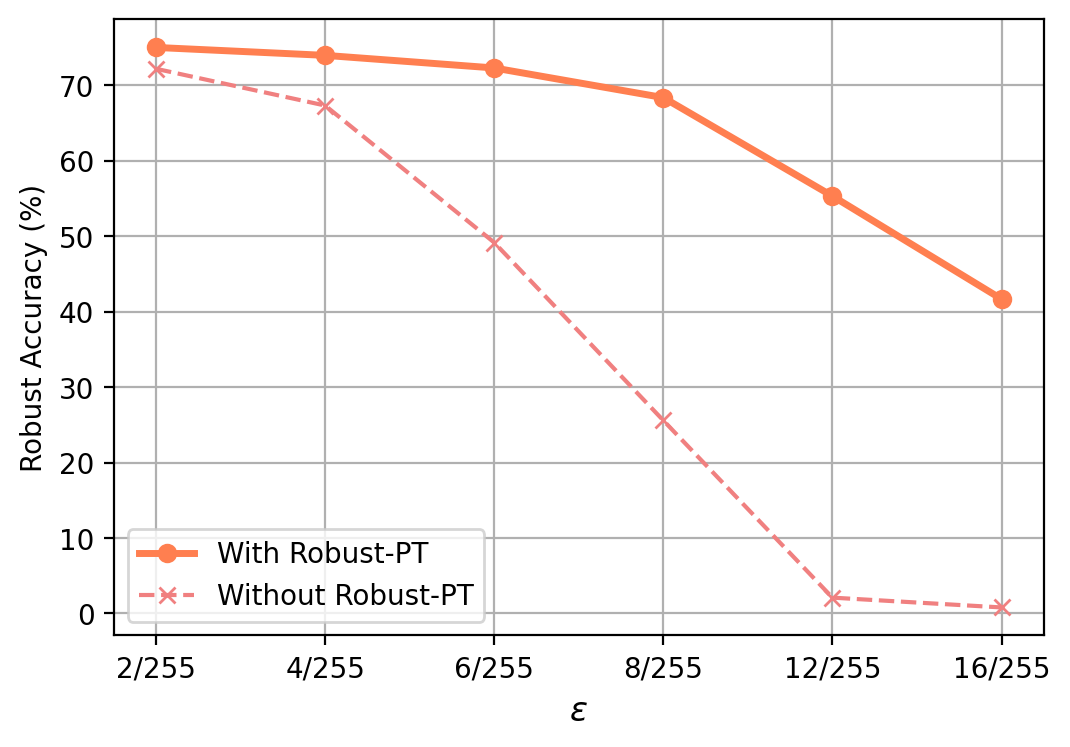}
        \caption{TAROT w/ and w/o R-PT}
        \label{fig:robust_tarot}
    \end{subfigure}
    \hfill
    \\
    \begin{subfigure}[t]{0.45\columnwidth}
        \centering
        \includegraphics[width=\textwidth]{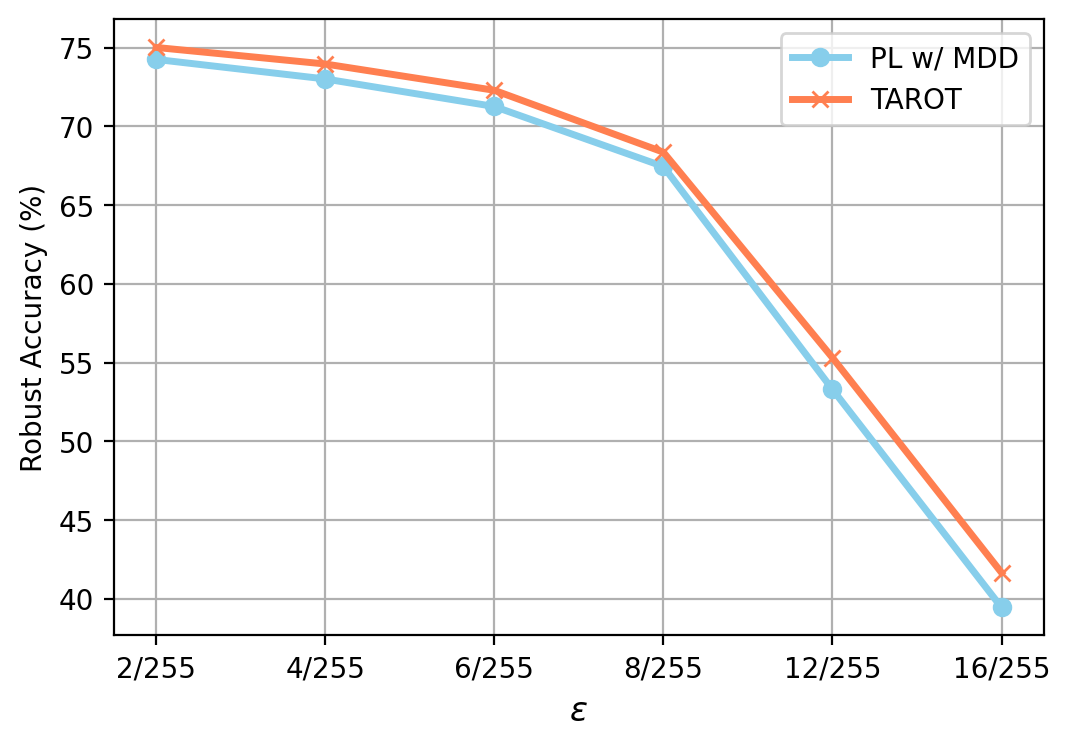}
        \caption{Rob Acc. w/ R-PT}
        \label{fig:robust_comparison}
    \end{subfigure} 
    \hfill
    \begin{subfigure}[t]{0.45\columnwidth}
        \centering
        \includegraphics[width=\textwidth]{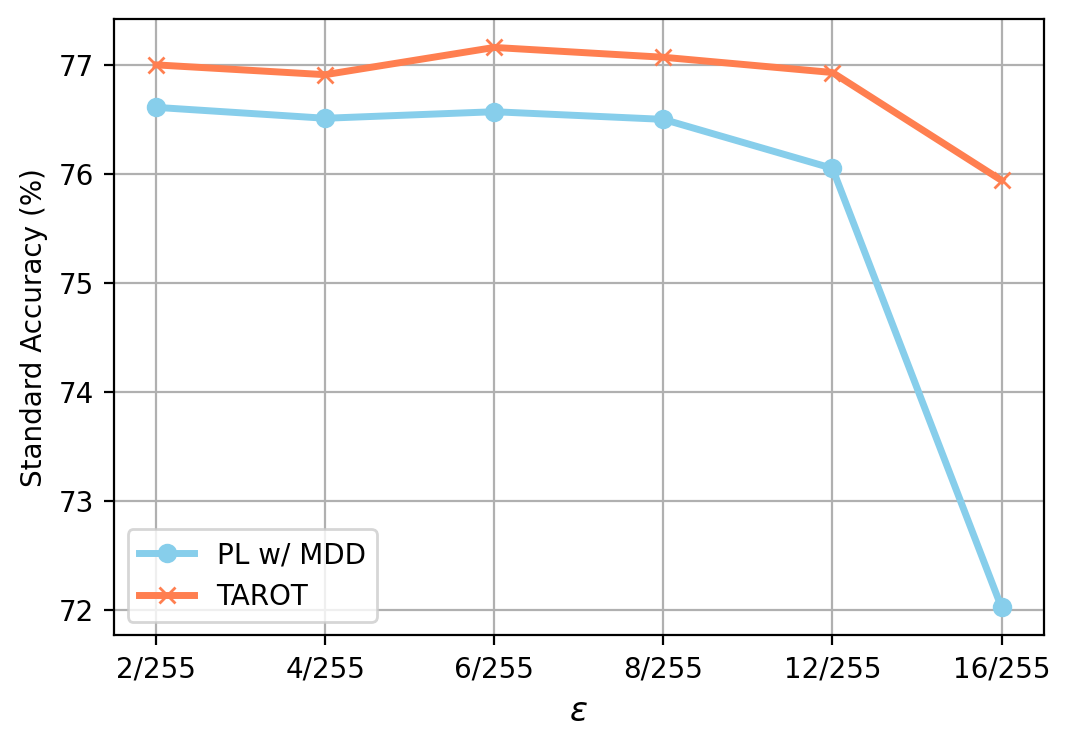}
        \caption{Stand Acc. w/ R-PT}
        \label{fig:standard_comparison}
    \end{subfigure}
    \caption{Effect of Robust-PT across various methods and $\varepsilon$ values. Each plot compares different configurations of PL w/ MDD and TAROT for standard and robust accuracy.}
    \label{fig:robust_pt_effect}
    \vskip -0.3in
\end{figure}

\paragraph{Effect of Robust-PT on Various $\varepsilon$}
\Cref{fig:robust_pl_mdd,fig:robust_tarot} show that 
the both algorithms perform significantly better when Robust-PT is applied, particularly as $\varepsilon$ increases, indicating that Robust-PT is crucial for robustness against strong attack.
Without Robust-PT, both TAROT and PL experience sharp declines in accuracy, especially as $\varepsilon$ becomes larger.
\Cref{fig:robust_comparison,fig:standard_comparison} illustrate that 
TAROT uniformly outperforms PL across different perturbation budgets ($\varepsilon$)
both on the standard accuracies and robust accuracies against AA.
Especially, that performance gap is increasing as $\varepsilon$ increases. %
\Cref{fig:robust_pl_mdd,fig:robust_tarot} show that both algorithms perform significantly better when Robust-PT is applied, particularly as  $\varepsilon$ increases, indicating that Robust-PT is crucial for robustness against strong attacks.
\Cref{fig:robust_comparison,fig:standard_comparison} illustrate that TAROT uniformly outperforms the PL approach across different perturbation budgets ($\varepsilon$), both in standard accuracies and robust accuracies against AA. 
Notably, the performance gap increases as $\varepsilon$ increases. 

\paragraph{Sensitivity Analysis of $\alpha$}
\Cref{fig:sensitivity_alpha} illustrates the sensitivity with respect to $\alpha$ in \cref{objective-tarot}. It shows how the performance of generalization and robustness changes for the target and unseen domains as the value of $\alpha$ varies.
As shown in the figure, each task has a different optimal $\alpha$, indicating that the optimal $\alpha$ value for one task may not be optimal for another.
The detailed numbers are provided in \cref{app:alpha}.
\section{Conclusions and Future Works}
\label{sec:conclusion}
To our best knowledge, this is the first work for adversarially robust domain adaptation algorithms grounded with theoretical explanations. 
Compared to existing works,
our algorithm shows essentially domain invariant robust features, resulting significant improved performance in source and unseen domains.
Our work not only serves as a cornerstone for understanding robust domain adaptation from a theoretical perspective, but we also believe TAROT can be served as a state-of-the-art algorithm in practical perspective by excelling researches of adversarial robustness for
transfer learning such as domain adaptation/generalization
and few-shot learning task.

{
    \paragraph{Acknowledgement}
    \small
    This work is partly supported by Institute of Information \& communications Technology Planning \& Evaluation (IITP) grant funded by the Korea government(MSIT) (No.RS-2022-II220184, Development and Study of AI Technologies to Inexpensively Conform to Evolving Policy on Ethics), National Research Foundation of Korea(NRF) grant funded by the Korea government(MSIT)(No. 2022R1A5A7083908) and Institute of Information \& communications Technology Planning \& Evaluation (IITP) grant funded by the Korea government(MSIT) [NO.RS-2021-II211343, Artificial Intelligence Graduate School Program (Seoul National University)].
    \bibliographystyle{ieeenat_fullname}
    \bibliography{main}
}

\clearpage
\setcounter{page}{1}
\maketitlesupplementary

\section{Novelty Summarization}
Our work goes beyond a simple theoretical extension of the existing MDD in three key aspects. 
\textbf{First}, we expand upon MDD by introducing a newly defined robust divergence to derive an upper bound
of the target domain robust risk that does not use  adversarial samples in the source domain.
Instead, the robust divergence measures the distance between distributions of 
clean examples in the source domain and adversarial examples in the target domain and thus
we can save computation times for generating adversarial samples in the source domain.
Moreover, an interesting property of TAROT is that the trained model is robust not only on the target domain but also on the source domain (see Proposition 3 in Section 7.3 in Appendix for theoretical evidence and Table 3 in Section 5.1.2 for empirical evidence) even if no adversarial samples in the source domain are used in the training phase.
Note that the replacing $\mathcal{R}_{\mathcal{S}}$ with $\mathcal{R} ^{\text{rob}}_{\mathcal{S}}$ also 
becomes an upper bound, but it is a looser bound than ours since $\mathcal{R}_{\mathcal{S}}(f) \leq \mathcal{R} ^{\text{rob}}_{\mathcal{S}}(f)$.
\textbf{Second}, the local Lipschitz constant, a newly introduced term, provides a theoretical bridge to existing research on adversarial robustness \cite{yang2020closer, zhang2022rethinking}. 
\textbf{Finally}, by offering a partial theoretical explanation of the existing robust UDA algorithms, 
we integrate the previous works and pave a new direction for robust UDA, highlighting novel approaches and potential advancements in the field. Hence we believe that our bound 
is cleverly devised for robust UDA beyond a simple extenstion of MDD.

\section{Theoretical Results}
\label{sec:theo_res}
\subsection{Auxiliary Lemmas}
\label{subsec:aux_lem}
\begin{lemma}[Lemma C.4 from \citet{zhang2019bridging},  Theorem 8.1 from \citet{mohri2012foundation}]
\label{lem:gen_bound}
    Let $\mathcal{F}\subseteq\mathbb{R}^{\mathcal{X}\times\mathcal{Y}}$ be a hypothesis set of score functions where $\mathcal{Y}=\{1,\dots,C\}$. 
    Define 
    \begin{equation*}
        \Pi_1\mathcal{F}=\{ \bm{x}\mapsto f(\bm{x},y)\vert y\in\mathcal{Y},f\in\mathcal{F} \}
    \end{equation*}
    and fix the margin parameter $\rho>0$. 
    Then for any $\delta>0$, with probability at least $1-\delta$, the following inequality holds for all $f\in\mathcal{F}$. 
    \begin{equation*}
        |\mathcal{R}_\mathcal{D}^{(\rho)}(f)-\mathcal{R}_{\widehat{\mathcal{D}}}^{(\rho)}(f)|\le\frac{2C^2}{\rho}\mathfrak{R}_{n,\mathcal{D}}(\Pi_1\mathcal{F})+\sqrt{\frac{\log\frac{2}{\delta}}{2n}}
    \end{equation*}
\end{lemma}


\begin{lemma}[Talagrand's lemma \cite{talagrand2014upper,mohri2012foundation}]
\label{lem:talagrand}
    Let $\Phi:\mathbb{R}\to\mathbb{R}$ be an $l$-Lipschitz function. Then for any hypothesis set $\mathcal{H}$ of real-valued functions and any samples $\widehat{\mathcal{D}}$ of size $n$, the following inequality holds: 
    \begin{equation*}
        \widehat{\mathfrak{R}}_{\widehat{\mathcal{D}}}(\Phi\circ H)\le l\widehat{\mathfrak{R}}_{\widehat{\mathcal{D}}}(H)
    \end{equation*}
\end{lemma}

\begin{lemma}[Lemma 8.1 from \citet{mohri2012foundation}]
\label{lem:rademacher_sum}
    Consider $k>1$ hypothesis sets $\mathcal{F}_1,\dots,\mathcal{F}_k$ in $\mathbb{R}^\mathcal{X}$. 
    Let $\mathcal{G}=\{\max\{h_1,\dots,h_l\}:h_i\in\mathcal{F}_j,j\in[1,l]\}$. 
    Then for any sample $\widehat{\mathcal{D}}$ size of $n$, the following holds: 
    \begin{equation}
        \widehat{\mathfrak{R}}_{\widehat{\mathcal{D}}}(\mathcal{G})\le\sum_{j=1}^l\widehat{\mathfrak{R}}_{\widehat{\mathcal{D}}}(\mathcal{F}_j)
    \end{equation}
\end{lemma}

\subsection{Proofs}
\label{subsec:proofs}



\begin{lemma}
\label{lem:disp_to_risks}
    For any distribution $\mathcal{D}$ which $(\mathbf{X},Y)$ follows, and score functions $f,f'$, the following inequality holds. 
    \begin{equation}
        \operatorname{disp}^{(\rho)}_{\mathcal{D}_\mathbf{X}}(f',f)\le\mathcal{R}^{(\rho)}_\mathcal{D}(f')+\mathcal{R}^{(\rho)}_\mathcal{D}(f)
    \end{equation}
\end{lemma}

\begin{proof}
    We first show that the following holds.
    \begin{equation*}
        \Phi_\rho\circ\cm_f(\mathbf{X},h_{f'}(\mathbf{X}))\le\Phi_\rho\circ\cm_{f'}(\mathbf{X},Y)+\Phi_\rho\circ\cm_f(\mathbf{X},Y)
    \end{equation*}
    
    If $h_{f'}(\mathbf{X})\ne Y$ or $h_{f}(\mathbf{X})\ne Y$ holds, then the right-hand side is bigger than $1$, consequently the inequality holds. 
    Now consider the case $h_{f'}(\mathbf{X})=h_{f}(\mathbf{X})=Y$. 
    Since $\Phi_\rho\circ\cm_f(\mathbf{X},h_f(\mathbf{X}))=\Phi_\rho\circ\cm_f(\mathbf{X},Y)$ holds, the wanted inequality holds. 

    Therefore, following inequality holds. 
    \begin{align*}
        &\operatorname{disp}_{\mathcal{D}_\mathbf{X}}^{(\rho)}(f',f)\\
        &=\mathbb{E}_{\mathbf{X}\sim\mathcal{D}}\Phi_\rho\circ\cm_f(\mathbf{X},h_{f'}(\mathbf{X}))\\
        &\le\mathbb{E}_{\mathbf{X}\sim\mathcal{D}}\Phi_\rho\circ\cm_{f'}(\mathbf{X},Y)+\mathbb{E}_{\mathbf{X}\sim\mathcal{D}}\Phi_\rho\circ\cm_f(\mathbf{X},Y)\\
        &=\mathcal{R}_\mathcal{D}^{(\rho)}(f')+\mathcal{R}_\mathcal{D}^{(\rho)}(f)
    \end{align*}
\end{proof}
    


\tarotprop*

\begin{proof}
Since 
\begin{align*}
    &\advmax\mathbbm{1}\{ Y\ne h_f(\mathbf{X}') \}\\
    &\le\advmax\mathbbm{1}\{ h_{f^{*}}(\mathbf{X}) \ne h_f(\mathbf{X}') \} + \mathbbm{1}\{ h_{f^{*}}(\mathbf{X}) \neq Y\}
\end{align*}
holds, we can start to derive the following inequalities.
\begin{align*}
    &\mathcal{R}^{\text{rob}}_{\mathcal{T}}(f) \\
    &\le\operatorname{disp}^{\text{rob}}_{\mathcal{T}_{\mathbf{X}}}(f^{*}, f) + \mathcal{R}_{\mathcal{T}}(f^{*}) \\
    &\le\operatorname{disp}^{\text{rob},(\rho)}_{\mathcal{T}_{\mathbf{X}}}(f^{*},f)+\mathcal{R}_\mathcal{T}^{(\rho)}(f^*)\\
    &=\operatorname{disp}^{\text{rob},(\rho)}_{\mathcal{T}_\mathbf{X}}(f^*,f)+\mathcal{R}_\mathcal{T}^{(\rho)}(f^*)+\mathcal{R}_\mathcal{S}^{(\rho)}(f)-\mathcal{R}_\mathcal{S}^{(\rho)}(f)\\
    &\le\operatorname{disp}^{\text{rob},(\rho)}_{\mathcal{T}_\mathbf{X}}(f^*,f)+\mathcal{R}_\mathcal{T}^{(\rho)}(f^*)+\mathcal{R}_\mathcal{S}^{(\rho)}(f)\\
    &\quad+\mathcal{R}_\mathcal{S}^{(\rho)}(f^*)-\operatorname{disp}^{(\rho)}_{\mathcal{S}_\mathbf{X}}(f^*,f)
    \\
    &=\mathcal{R}_\mathcal{S}^{(\rho)}(f)+ \operatorname{disp}^{\text{rob},(\rho)}_{\mathcal{T}_\mathbf{X}}(f^*,f)-\operatorname{disp}^{(\rho)}_{\mathcal{S}_\mathbf{X}}(f^*,f) +\lambda
\end{align*}
Here, the third inequality holds from \cref{lem:disp_to_risks} and $\lambda=\mathcal{R}_\mathcal{T}^{(\rho)}(f^*)+\mathcal{R}_\mathcal{S}^{(\rho)}(f^*)$. 
\end{proof}

\begin{lemma}[Part of Theorem C.7 from \citet{zhang2019bridging}]
\label{lem:disp_diff_standard}
    For a given distribution $\mathcal{D}$, corresponding empirical distribution $\widehat{\mathcal{D}}$, and any $\delta>0$, with probability at least $1-\delta$, the following holds for $\forall f,f'\in\mathcal{F}$ simultaneously. 
    \begin{equation}
    \begin{aligned}
        &\left\lvert \operatorname{disp}_{\mathcal{D}_\mathbf{X}}^{(\rho)}(f',f)-\operatorname{disp}_{\widehat{\mathcal{D}}_\mathbf{X}}^{(\rho)}(f',f) \right\rvert\\
        &\le\frac{2C}{\rho}\mathfrak{R}_{n,\mathcal{D}}(\Pi_\mathcal{H}\mathcal{F})+\sqrt{\frac{\log\frac{2}{\delta}}{2n}}
    \end{aligned}
    \end{equation}
\end{lemma}

\begin{lemma}
\label{lem:disp_diff_robust}
    For a given distribution $\mathcal{D}$, its marginal distribution $\mathcal{D}_\mathbf{X}$, corresponding empirical distribution $\widehat{\mathcal{D}}_\mathbf{X}$, and any $\delta>0$, with probability at least $1-\delta$, the following holds for $\forall f,f'\in\mathcal{F}$ simultaneously. 
    \begin{equation}
    \begin{aligned}
        &\left\lvert \operatorname{disp}_{\mathcal{D}_\mathbf{X}}^{\text{rob},(\rho)}(f',f)-\operatorname{disp}_{\widehat{\mathcal{D}}_\mathbf{X}}^{\text{rob},(\rho)}(f',f) \right\rvert\\
        &\le\frac{2C}{\rho}\mathfrak{R}_{n,\mathcal{D}}(\Pi_\mathcal{H}\mathcal{F})+\sqrt{\frac{\log\frac{2}{\delta}}{2n}}+\frac{2\varepsilon L_{f}(\mathcal{D}_\mathbf{X},\varepsilon)}{\rho}
    \end{aligned}
    \end{equation}
\end{lemma}
\begin{proof}
Denote $f(\bm{x})=(f(\bm{x},1),\dots,f(\bm{x},C))^T\in\mathbb{R}^C$ for arbitrary $\bm{x}\in\mathcal{X}$. 
Note that $\Phi_\rho$ is $1/\rho$-Lipschitz, and the margin operator is $2$-Lipschitz \cite{bartlett2006convexity}. 
Hence, for $\forall\bm{x}'\in\mathcal{B}_p(\bm{x},\varepsilon)$, the following inequality holds. 
\begin{align*}
    &\left\lvert \Phi_\rho\circ\cm_f(\bm{x}',y)-\Phi_\rho\circ\cm_f(\bm{x},y) \right\rvert\\
    &\le\frac{1}{\rho}\left\lvert \cm_f(\bm{x}',y)-\cm_f(\bm{x}',y) \right\rvert\\
    &\le\frac{2}{\rho}\left\lVert f(\bm{x}')-f(\bm{x}) \right\rVert_1\\
    &\le\frac{2}{\rho}\varepsilon L_f(\mathcal{D}_\mathbf{X},\varepsilon)
\end{align*}

Here we utilize the proof technique from \citet{zhang2019bridging}. 
For $\forall f,f'\in\mathcal{F}$, define the $\tau_{f'}$-transform of $f$ as follows: 
\begin{equation*}
    \tau_{f'}f(\bm{x},y)=\begin{cases}
        f(\bm{x},1) & \text{if }y=h_{f'}(\bm{x})\\
        f(\bm{x},h_{f'}(\bm{x})) & \text{if }y=1\\
        f(\bm{x},y) & \text{o.w.}
    \end{cases}
\end{equation*}
where $h_f$ is the induced classifier from $f$. 
Let $\mathcal{G}=\{\tau_{f'}f\vert f,f'\in\mathcal{F}\}$ and $\tilde{\mathcal{G}}=\{(x,y)\mapsto\cm_g(x,y)\vert g\in\mathcal
{G}\}$. 
Now using these sets, we can represent the disparity terms into risk terms. 
For any $f,f'\in\mathcal{F}$, let $g=\tau_{f'}f$. 

Then, 
\begin{align*}
    &\cm_g(\bm{x},1)\\
    &=\cm_{\tau_{f'}f}(\bm{x},1)\\
    &=\tau_{f'}f(\bm{x},1)-\underset{y'\ne 1}{\max}\tau_{f'}f(\bm{x},y')\\
    &=f(\bm{x},h_{f'}(\bm{x}))-\max\left\{\underset{y'\ne 1,h_{f'}(\bm{x})}{\max}f(\bm{x},y'),f(\bm{x},1)\right\}\\
    &=f(\bm{x},h_{f'}(\bm{x}))-\underset{y'\ne h_{f'}(\bm{x})}{\max}f(\bm{x},y)\\
    &=\cm_f(\bm{x},h_{f'}(\bm{x}))
\end{align*}
holds. 


Hence, 
\begin{align*}
    &\operatorname{disp}_{\mathcal{D}_\mathbf{X}}^{\text{rob},(\rho)}(f',f)\\
    &=\mathbb{E}_{\mathbf{X}\sim\mathcal{D}_\mathbf{X}}\advmax\Phi_\rho\circ\cm_f(\mathbf{X}',h_{f'}(\mathbf{X}))\\
    &\le\mathbb{E}_{\mathbf{X}\sim\mathcal{D}_\mathbf{X}}\Phi_\rho\circ\cm_f(\mathbf{X},h_{f'}(\mathbf{X}))+\frac{2\varepsilon L_{f}(\mathcal{D}_\mathbf{X},\varepsilon)}{\rho}\\
    &=\mathbb{E}_{\mathbf{X}\sim\mathcal{D}_\mathbf{X}}\Phi_\rho\circ\cm_g(\mathbf{X},1)+\frac{2\varepsilon L_{f}(\mathcal{D}_\mathbf{X},\varepsilon)}{\rho}
\end{align*}
holds for $g=\tau_{f'}f$. 

For arbitrary set of score functions $\mathcal{U}$, we define following term: 
\begin{equation*}
    \mathfrak{R}_{n,\mathcal{D}}^0(\mathcal{U}):=\mathbb{E}_{(\bm{x}_i,1),\bm{x}_i\sim \mathcal{D}^n}\hat{\mathfrak{R}}_{\widehat{\mathcal{D}}}(\mathcal{U})
\end{equation*}

Regard all the data as from the same class $1$.
Then by using \cref{lem:gen_bound}, the following inequality holds simultaneously for any $g\in\mathcal{G}$, with probability at least $1-\delta$, 
\begin{align*}
    &\mathbb{E}_{\mathbf{X}\sim\mathcal{D}_\mathbf{X}}\Phi_\rho\circ\cm_g(\mathbf{X},1)\\
    &\le\mathbb{E}_{\mathbf{X}\sim\widehat{\mathcal{D}}_\mathbf{X}}\Phi_\rho\circ\cm_g(\mathbf{X},1)+2\mathfrak{R}_{n,\mathcal{D}}^0(\Phi\circ\tilde{\mathcal{G}})+\sqrt{\frac{\log\frac{2}{\delta}}{2n}}
\end{align*}


Hence, 
\begin{align*}
    &\operatorname{disp}_{\mathcal{D}_\mathbf{X}}^{\text{rob},(\rho)}(f',f)\\
    &\le\mathbb{E}_{\mathbf{X}\sim\widehat{\mathcal{D}}}\Phi_\rho\circ\cm_g(\mathbf{X},1)+2\mathfrak{R}_{n,\mathcal{D}}^0(\Phi\circ\tilde{\mathcal{G}})\\
    &\quad+\sqrt{\frac{\log\frac{2}{\delta}}{2n}}+\frac{2\varepsilon L_{f}({\mathcal{D}}_\mathbf{X},\varepsilon)}{\rho}\\
    &=\mathbb{E}_{\mathbf{X}\sim\widehat{\mathcal{D}}}\Phi_\rho\circ\cm_{f'}(\mathbf{X},h_f(\mathbf{X}))+2\mathfrak{R}_{n,\mathcal{D}}^0(\Phi\circ\tilde{\mathcal{G}})]\\
    &\quad+\sqrt{\frac{\log\frac{2}{\delta}}{2n}}+\frac{2\varepsilon L_{f}({\mathcal{D}}_\mathbf{X},\varepsilon)}{\rho}\\
    &\le\mathbb{E}_{\mathbf{X}\sim\widehat{\mathcal{D}}}\advmax\Phi_\rho\circ\cm_{f'}(\mathbf{X}',h_f(\mathbf{X}))+2\mathfrak{R}_{n,\mathcal{D}}^0(\Phi\circ\tilde{\mathcal{G}})\\
    &\quad+\sqrt{\frac{\log\frac{2}{\delta}}{2n}}+\frac{2\varepsilon L_{f}({\mathcal{D}}_\mathbf{X},\varepsilon)}{\rho}\\
    &=\operatorname{disp}_{\widehat{\mathcal{D}}_\mathbf{X}}^{\text{rob},(\rho)}(f',f)+2\mathfrak{R}_{n,\mathcal{D}}^0(\Phi\circ\tilde{\mathcal{G}})\\
    &\quad+\sqrt{\frac{\log\frac{2}{\delta}}{2n}}+\frac{2\varepsilon L_{f}({\mathcal{D}}_\mathbf{X},\varepsilon)}{\rho}
\end{align*}
holds. 
Now, we want to bound the term $\mathfrak{R}_{n,\mathcal{D}}^0(\Phi\circ\tilde{\mathcal{G}})$. 


By \Cref{lem:talagrand}, 
\begin{equation*}
    \mathfrak{R}_{n,\mathcal{D}}^0(\Phi\circ\tilde{\mathcal{G}})\le\frac{1}{\rho}\mathfrak{R}_{n,\mathcal{D}}^0(\tilde{\mathcal{G}})
\end{equation*}
holds. 
Also, 
\begin{align*}
    &\mathfrak{R}_{n,\mathcal{D}}^0(\tilde{\mathcal{G}})\\
    &=\frac{1}{n}\expD\underset{g\in\mathcal{G}}{\sup}\sum_{i=1}^n\sigma_i\cm_g(\bm{x}_i,1)\\
    &\le\frac{1}{n}\expD\underset{f\in\mathcal{F},h\in\mathcal{H}}{\sup}\sum_{i=1}^n\sigma_i f(\bm{x}_i,h(\bm{x}_i))\\
    &\quad+\frac{1}{n}\expD\underset{f\in\mathcal{F},h\in\mathcal{H}}{\sup}\sum_{i=1}^n\sigma_i\left(-\underset{y\ne h(\bm{x}_i)}{\max}f(\bm{x}_i,y)\right)\\
    &=\mathfrak{R}_{n,\mathcal{D}}(\Pi_\mathcal{H}\mathcal{F})+\frac{1}{n}\expD\underset{f\in\mathcal{F},h\in\mathcal{H}}{\sup}\sum_{i=1}^n\sigma_i\underset{y\ne h(\bm{x}_i)}{\max}f(\bm{x}_i,y)
\end{align*}
holds. 

Define the permutation
\begin{equation*}
    \xi(i)=\begin{cases}
        i+1 & i=1,\dots,C-1\\
        1 & i=C
    \end{cases}
\end{equation*}

As we assumed that $\mathcal{H}$ is permutation-invariant, we know that for $\forall h\in\mathcal{H}$ and $j=1,\dots,k-1$, $\xi^jh\in\mathcal{H}$ holds. 
Let $\Pi_\mathcal{H}\mathcal{F}^{(C-1)}=\{\max\{f_1,\dots,f_{C-1}\}\vert f_i\in\Pi_\mathcal{H}\mathcal{F},i=1,\dots,C-1\}$. 

Then, 
\begin{align*}
    &\frac{1}{n}\expD\underset{f\in\mathcal{F},h\in\mathcal{H}}{\sup}\sum_{i=1}^n\sigma_i\underset{y\ne h(\bm{x}_i)}{\max}f(\bm{x}_i,y)\\
    &=\frac{1}{n}\expD\underset{f\in\mathcal{F},h\in\mathcal{H}}{\sup}\sum_{i=1}^n\sigma_i\underset{j\in\{1,\dots,k-1\}}{\max}f(\bm{x}_i,\xi^jh(\bm{x}_i))\\
    &=\frac{1}{n}\expD\underset{f\in\Pi_\mathcal{H}\mathcal{F}^{(C-1)}}{\sup}\sum_{i=1}^n\sigma_i f(\bm{x}_i)\\
    &\le\frac{C-1}{n}\expD\underset{f\in\Pi_\mathcal{H}\mathcal{F}}{\sup}\sum_{i=1}^n\sigma_i f(\bm{x}_i)
\end{align*}
holds, where the last inequality holds from \Cref{lem:rademacher_sum}. 
Hence, 
\begin{align*}
    &\mathfrak{R}_{n,\mathcal{D}}^0(\tilde{\mathcal{G}})\\
    &\le\mathfrak{R}_{n,\mathcal{D}}(\Pi_\mathcal{H}\mathcal{F})+\frac{C-1}{n}\expD\underset{f\in\Pi_\mathcal{H}\mathcal{F}}{\sup}\sum_{i=1}^n\sigma_i f(\bm{x}_i)\\
    &\le C\mathfrak{R}_{n,\mathcal{D}}(\Pi_\mathcal{H}\mathcal{F})
\end{align*}
holds. 

Combining above inequalities, we have the following inequality  
\begin{align*}
    &\operatorname{disp}_{\mathcal{D}_\mathbf{X}}^{\text{rob},(\rho)}(f',f)\\
    &\le\operatorname{disp}_{\widehat{\mathcal{D}}_\mathbf{X}}^{\text{rob},(\rho)}(f',f)+\frac{2C}{\rho}\mathfrak{R}_{n,\mathcal{D}}(\Pi_\mathcal{H}\mathcal{F})\\
    &\quad+\sqrt{\frac{\log\frac{2}{\delta}}{2n}}+\frac{2\varepsilon L_{f}(\mathcal{D}_\mathbf{X},\varepsilon)}{\rho}
\end{align*}
holds simultaneously for $\forall f,f'\in\mathcal{F}$ with probability at least $1-\delta$. 

In the same way, we have the opposite direction by exchanging $\mathcal{D}$ and $\widehat{\mathcal{D}}$. 
Therefore, the following holds simultaneously for $\forall f,f'\in\mathcal{F}$ with probability at least $1-\delta$, 
\begin{align*}
    &\left\lvert \operatorname{disp}_{\mathcal{D}_\mathbf{X}}^{\text{rob},(\rho)}(f',f)-\operatorname{disp}_{\widehat{\mathcal{D}}_\mathbf{X}}^{\text{rob},(\rho)}(f',f) \right\rvert\\
    &\le\frac{2C}{\rho}\mathfrak{R}_{n,\mathcal{D}}(\Pi_\mathcal{H}\mathcal{F})+\sqrt{\frac{\log\frac{2}{\delta}}{2n}}+\frac{2\varepsilon L_{f}(\mathcal{D}_\mathbf{X},\varepsilon)}{\rho}
\end{align*}
concluding the proof. 

\end{proof}

\begin{restatable}{lemma}{robustemp}
\label{lem3}
For any $\delta>0$, with probability at least $1-\delta$, the following holds for all $f\in\mathcal{F}$.
\begin{equation}
\begin{aligned}
    &\lvert\mathcal{R}_\mathcal{D}^{\text{rob}, (\rho)}(f)-\mathcal{R}_{\widehat{\mathcal{D}}}^{\text{rob}, (\rho)}(f)\rvert\\
    &\le\frac{2C^2}{\rho}\mathfrak{R}_{n,\mathcal{D}}(\Pi_1\mathcal{F})+\sqrt{\frac{\log\frac{2}{\delta}}{2n}}+\frac{2\varepsilon L_f(\mathcal{X},\varepsilon)}{\rho}
\end{aligned}
\end{equation}
\end{restatable}

\begin{proof}

We know that
\begin{equation*}
    \lvert \Phi_\rho\circ\cm_f(\bm{x}', y)-\Phi_\rho\circ\cm_f(\bm{x}, y) \rvert\le\frac{2}{\rho}\varepsilon L_f(\mathcal{D}_\mathbf{X},\varepsilon)
\end{equation*}
holds for $\forall \bm{x}'\in\mathcal{B}_p(\bm{x},\varepsilon)$. 

Then, the following holds with probability at least $1-\delta$.
\begin{align*}
&\mathcal{R}_\mathcal{D}^{\text{rob}, (\rho)}(f)\\
&=\mathbb{E}_\mathcal{D}\max_{\mathbf{X}'\in\mathcal{B}_p(\mathbf{X},\varepsilon)}\Phi_\rho\circ\cm_f(\mathbf{X}',\mathbf{Y})\\
&\le\mathbb{E}_\mathcal{D}\Phi_\rho\circ\cm_f(\mathbf{X},\mathbf{Y})+\frac{2\varepsilon L_f(\mathcal{D}_\mathbf{X},\varepsilon)}{\rho}\\
&\le\mathcal{R}_{\widehat{\mathcal{D}}}^{(\rho)}(f)+\frac{2C^2}{\rho}\mathfrak{R}_{n,\mathcal{D}}(\Pi_1\mathcal{F})+\sqrt{\frac{\log\frac{2}{\delta}}{2n}}+\frac{2\varepsilon L_f(\mathcal{D}_\mathbf{X},\varepsilon)}{\rho}\\
&=\frac{1}{n}\sum_{i=1}^n\Phi_\rho\circ\cm_f(\bm{x}_i,y_i)+\frac{2C^2}{\rho}\mathfrak{R}_{n,\mathcal{D}}(\Pi_1\mathcal{F})\\
&\quad+\sqrt{\frac{\log\frac{2}{\delta}}{2n}}+\frac{2\varepsilon L_f(\mathcal{D}_\mathbf{X},\varepsilon)}{\rho}\\
&\le\frac{1}{n}\sum_{i=1}^n\max_{\bm{x}_i'\in\mathcal{B}_p(\bm{x}_i,\varepsilon)}\Phi_\rho\circ\cm_f(\bm{x}'_i, y_i)+\frac{2C^2}{\rho}\mathfrak{R}_{n,\mathcal{D}}(\Pi_1\mathcal{F})\\
&\quad+\sqrt{\frac{\log\frac{2}{\delta}}{2n}}+\frac{2\varepsilon L_f(\mathcal{D}_\mathbf{X},\varepsilon)}{\rho}\\
&=\mathcal{R}_{\widehat{\mathcal{D}}}^{\text{rob}, (\rho)}(f)+\frac{2C^2}{\rho}\mathfrak{R}_{n,\mathcal{D}}(\Pi_1\mathcal{F})+\sqrt{\frac{\log\frac{2}{\delta}}{2n}}+\frac{2\varepsilon L_f(\mathcal{D}_\mathbf{X},\varepsilon)}{\rho}
\end{align*}
\end{proof}

\begin{restatable}{lemma}{robustmddemp}
\label{lem:d_rob_bound}
For any $\delta$ > 0, with probability $1-2\delta$, the following holds simultaneously for any score function $f$, 

\begin{equation}
\begin{aligned}
    &\left\lvert d_{f, \mathcal{F}}^{\text{rob}, (\rho)}(\widehat{\mathcal{S}}_\mathbf{X}, \widehat{\mathcal{T}}_\mathbf{X})-d_{f, \mathcal{F}}^{\text{rob}, (\rho)}(\mathcal{S}_\mathbf{X}, \mathcal{T}_\mathbf{X})\right\rvert\\
    &\leq\frac{2C}{\rho} \mathfrak{R}_{n, \mathcal{S}}\left(\Pi_{\mathcal{H}} \mathcal{F}\right)+\frac{2k}{\rho}\mathfrak{R}_{m, \mathcal{T}}\left(\Pi_{\mathcal{H}} \mathcal{F}\right)\\
    &\quad+\sqrt{\frac{\log \frac{2}{\delta}}{2n}}+\sqrt{\frac{\log \frac{2}{\delta}}{2m}}+\frac{2\varepsilon L_f(\mathcal{T}_\mathbf{X},\varepsilon)}{\rho}
\end{aligned}
\end{equation}
\end{restatable}

\begin{proof}

From \Cref{lem:disp_diff_robust}, we have 
\begin{align*}
    &\left\lvert\operatorname{disp}_{\mathcal{T}_\mathbf{X}}^{\text{rob},(\rho)}(f',f)-\operatorname{disp}_{\widehat{\mathcal{T}}_\mathbf{X}}^{\text{rob},(\rho)}(f',f)\right\rvert\\
    &\le \frac{2C}{\rho}\mathfrak{R}_{n,\mathcal{T}}(\Pi_\mathcal{H}\mathcal{F})+\sqrt{\frac{\log\frac{2}{\delta}}{2n}}+\frac{2\varepsilon L_f(\mathcal{T}_\mathbf{X},\varepsilon)}{\rho}
\end{align*}
Also, from \Cref{lem:disp_diff_standard}, the following holds with probability at least $1-\delta$, 
\begin{align*}
    &\left\lvert\operatorname{disp}_{\mathcal{S}_\mathbf{X}}^{(\rho)}(f',f)-\operatorname{disp}_{\widehat{\mathcal{S}}_\mathbf{X}}^{(\rho)}(f',f)\right\rvert\\
    &\le \frac{2C}{\rho}\mathfrak{R}_{m,\mathcal{S}}(\Pi_\mathcal{H}\mathcal{F})+\sqrt{\frac{\log\frac{2}{\delta}}{2m}}
\end{align*}

Hence, 
\begin{align*}
    &\left\lvert d_{f,\mathcal{F}}^{\text{rob},(\rho)}(\mathcal{S}_\mathbf{X}, \mathcal{T}_\mathbf{X})-d_{f,\mathcal{F}}^{\text{rob},(\rho)}(\widehat{\mathcal{S}}_\mathbf{X}, \widehat{\mathcal{T}}_\mathbf{X})\right\rvert\\
    &=\Bigg\lvert\underset{f'\in\mathcal{F}}{\sup}\left\{ \operatorname{disp}_{\mathcal{T}_\mathbf{X}}^{\text{rob},(\rho)}(f',f)-\operatorname{disp}_{\mathcal{S}_\mathbf{X}}^{(\rho)}(f',f) \right\}\\
    &\quad-\underset{f'\in\mathcal{F}}{\sup}\left\{ \operatorname{disp}_{\widehat{\mathcal{T}}_\mathbf{X}}^{\text{rob},(\rho)}(f',f)-\operatorname{disp}_{\widehat{\mathcal{S}}_\mathbf{X}}^{(\rho)}(f',f) \right\}\Bigg\rvert\\
    &\le \sup_{f'\in\mathcal{F}}\left\lvert\operatorname{disp}_{\mathcal{S}_\mathbf{X}}^{(\rho)}(f',f)-\operatorname{disp}_{\widehat{\mathcal{S}}_\mathbf{X}}^{(\rho)}(f',f)\right\rvert\\
    &\quad+ \sup_{f'\in\mathcal{F}}\left\lvert\operatorname{disp}_{\mathcal{T}_\mathbf{X}}^{\text{rob},(\rho)}(f',f)-\operatorname{disp}_{\widehat{\mathcal{T}}_\mathbf{X}}^{\text{rob},(\rho)}(f',f)\right\rvert
\end{align*}
holds, concluding the proof.
\end{proof}

\tarotgb*
\begin{proof}
From \Cref{eq:upper_bound}, 
\begin{align*}
    &\mathcal{R}^{\text{rob}}_{\mathcal{T}}(h_f)\\
    & \le \mathcal{R}^{(\rho)}_{\mathcal{S}}(f) + \text{d}^{\text{rob}, (\rho)}_{f, \mathcal{F}}(\mathcal{S}_\mathbf{X}, \mathcal{T}_\mathbf{X}) + \lambda \\
    & \le \mathcal{R}_{\widehat{\mathcal{S}}}^{(\rho)}(f) + \frac{2C^2}{\rho}\mathfrak{R}_{n,\mathcal{S}}(\Pi_1\mathcal{F})+\sqrt{\frac{\log\frac{2}{\delta}}{2n}}\\
    &\quad+ \text{d}^{\text{rob}, (\rho)}_{f, \mathcal{F}}(\mathcal{S}_\mathbf{X}, \mathcal{T}_\mathbf{X}) + \lambda\\
    & \le \mathcal{R}_{\widehat{\mathcal{S}}}^{(\rho)}(f) + \frac{2C^2}{\rho}\mathfrak{R}_{n,\mathcal{S}}(\Pi_1\mathcal{F})+\sqrt{\frac{\log\frac{2}{\delta}}{2n}}\\
    &\quad + \text{d}^{\text{rob}, (\rho)}_{f, \mathcal{F}}(\widehat{\mathcal{S}}_\mathbf{X}, \widehat{\mathcal{T}}_\mathbf{X}) + \frac{2 C}{\rho} \mathfrak{R}_{n, \mathcal{S}}\left(\Pi_{\mathcal{H}} \mathcal{F}\right)\\
    &\quad+\frac{2 C}{\rho} \mathfrak{R}_{m, \mathcal{T}}\left(\Pi_{\mathcal{H}} \mathcal{F}\right)+\sqrt{\frac{\log \frac{2}{\delta}}{2 n}}+\sqrt{\frac{\log \frac{2}{\delta}}{2 m}}\\
    &\quad+\frac{2\varepsilon L_f(\mathcal{T}_\mathbf{X},\varepsilon)}{\rho} + \lambda \\ 
    & = \mathcal{R}_{\widehat{\mathcal{S}}}^{\text{rob}, (\rho)}(f) + \text{d}^{\text{rob}, (\rho)}_{f, \mathcal{F}}(\widehat{\mathcal{S}}, \widehat{\mathcal{T}}) + \lambda \\
    &\quad + \frac{2C^2}{\rho}\mathfrak{R}_{n,\mathcal{S}}(\Pi_1\mathcal{F}) +  \frac{2 C}{\rho} \mathfrak{R}_{n, \mathcal{S}}\left(\Pi_{\mathcal{H}} \mathcal{F}\right) \\
    &\quad+ \frac{2 C}{\rho} \mathfrak{R}_{m, \mathcal{T}}\left(\Pi_{\mathcal{H}} \mathcal{F}\right) + 2\sqrt{\frac{\log \frac{2}{\delta}}{2 n}}+\sqrt{\frac{\log \frac{2}{\delta}}{2 m}}\\
    &\quad+ \frac{2\varepsilon L_f(\mathcal{T}_\mathbf{X},\varepsilon)}{\rho}
\end{align*}
Here, the second inequality holds from \Cref{lem:gen_bound} and the third inequality holds from \Cref{lem:d_rob_bound}. 
\end{proof}

\subsection{Source Robusk Risk of TAROT}
\label{subsec:source_rob_risk_theo}
In this section, we derive an upper bound for the robust risk on the source domain. 
The components of following upper bound — standard source risk and robust disparity — correspond to the upper bound in \Cref{prop:upper_bound}, suggesting that our algorithm can effectively improve adversarial robustness on the source domain. 

\begin{proposition}
Consider a source domain $\mathcal{S}$, a target domain $\mathcal{T}$ and their marginal distributions $\mathcal{S}_\mathbf{X},\mathcal{T}_\mathbf{X}$ on $\mathbf{X}$ . 
    For every score function $f\in\mathcal{F}$, the following inequality holds: 
    \begin{equation}
        \mathcal{R}_\mathcal{S}^{\text{rob}}(f)\le\mathcal{R}_\mathcal{S}^{(\rho)}(f)+2d_{f,\mathcal{F}}^{\text{rob},(\rho)}(\mathcal{S}_\mathbf{X},\mathcal{T}_\mathbf{X})+\frac{2\varepsilon L_f(\mathcal{S}_\mathbf{X},\varepsilon)}{\rho}+\lambda
    \end{equation}
    where $\lambda=\underset{f\in\mathcal{F}}{\min}\{\mathcal{R}_\mathcal{T}^{(\rho)}(f)+\mathcal{R}_\mathcal{S}^{(\rho)}(f)\}$. 
\end{proposition}
\begin{proof}
    \begin{align*}
        &\mathcal{R}_\mathcal{S}^{\text{rob}}(f)\\
        &\le\mathcal{R}_\mathcal{T}^{(\rho)}(f)+\operatorname{disp}_{\mathcal{S}_\mathbf{X}}^{\text{rob},(\rho)}(f^*,f)-\operatorname{disp}_{\mathcal{T}_\mathbf{X}}^{(\rho)}(f^*,f)+\lambda\\
        &\le\mathcal{R}_\mathcal{T}^{(\rho)}(f)+\operatorname{disp}_{\mathcal{S}_\mathbf{X}}^{(\rho)}(f^*,f)-\operatorname{disp}_{\mathcal{T}_\mathbf{X}}^{(\rho)}(f^*,f)\\
        &\quad+\frac{2\varepsilon L_f(\mathcal{S}_\mathbf{X},\varepsilon)}{\rho}+\lambda\\
        &\le\mathcal{R}_\mathcal{S}^{(\rho)}(f)+2\operatorname{disp}_{\mathcal{S}_\mathbf{X}}^{(\rho)}(f^*,f)-\operatorname{disp}_{\mathcal{T}_\mathbf{X}}^{(\rho)}(f^*,f)\\
        &\quad+\frac{2\varepsilon L_f(\mathcal{S}_\mathbf{X},\varepsilon)}{\rho}+\lambda\\
        &\le\mathcal{R}_\mathcal{S}^{(\rho)}(f)+2d_{f,\mathcal{F}}^{\text{rob},(\rho)}(\mathcal{S}_\mathbf{X},\mathcal{T}_\mathbf{X})+\frac{2\varepsilon L_f(\mathcal{S}_\mathbf{X},\varepsilon)}{\rho}+\lambda
    \end{align*}
    Here, the first inequality holds by replacing $\mathcal{S}$ and $\mathcal{T}$ from \Cref{prop:upper_bound}. 
\end{proof}
Note that since the upper-bound considers the local Lipschitz constant on the source domain, this is a partial explanation for the source robust risk. 

\section{Further Details on Experiments}
\label{app:details}

\paragraph{Loss}
The exact forms of \Cref{objective-tarot}
loss function are as follows:
\begin{equation*}
    \ell_{\text{ce}} \left( ( \pi \circ \psi)(\bm{x}), y) \right) := - \log \sigma_y(\pi \circ \psi (\bm{x})),
\end{equation*}
\begin{equation*}
    \ell^{\text{rob}}_{\text{mod-ce}} \left( ( \pi \circ \psi)(\bm{x}), y) \right) := \log (1-\sigma_y(\pi \circ \psi (\bm{x}^{\text{adv}}))),
\end{equation*}
\begin{equation*}
    \ell^{\text{rob}}_{\text{ce}} \left( ( \pi \circ \psi)(\bm{x}), y) \right) := -\log \sigma_y(\pi \circ \psi (\bm{x}^{\text{adv}})),
\end{equation*}
where 
$\sigma_y$ denotes the predictive confidence for class 
$y$, i.e., the $y$-th component of the softmax output
and $\bm{x}^{\text{adv}}$ is the adversarial example.

\paragraph{Datasets}
Office-31 consists of 4,110 images from three domains — Amazon (A), Webcam (W), and DSLR (D) — 
considered to be classical data for domain adaptation due to the differences in image quality and capture methods.
Office-Home is more diverse, with 15,588 images across four domains — Art (Ar), Clipart (Cl), Product (Pr), and Realworld (Rw) — covering different styles, from artistic drawings to real photos.
VisDA2017 features over 280,000 images, focusing on the domain gap between synthetic and real images, providing a challenge for algorithms to handle synthetic (S) to real (R) adaptation.
DomainNet is the largest and challenging dataset, containing around 600,000 images from six domains, including Clipart (C), Infograph (I),  Sketch (S), Painting (P), Quickdraw (Q) and Real (R). 

\paragraph{Hyperparameters}
We follow the default experimental settings of \textit{TLlib} \cite{tllib}. 
We conduct experiments using the following training configuration. Models are trained for 20 epochs with a weight decay of \( 5 \times 10^{-4} \). Robust pretraining is set at \( \varepsilon = \frac{1}{255} \) for TAROT, PL, ARTUDA, and SRoUDA, 
while RFA uses models trained with different \( \varepsilon \) values identical to evaluation $\varepsilon$, as it does not directly generate adversarial examples during training.
We conduct experiments using the following training configuration. Models are trained for 20 epochs with a weight decay of 
\( 5 \times 10^{-4} \).
Robust pretraining is set at 
\( \varepsilon = \frac{1}{255} \) for TAROT, PL, ARTUDA, and SRoUDA. 
In contrast, RFA utilizes models trained with different 
$\varepsilon$ values matching the evaluation $\varepsilon$, as it does not directly generate adversarial examples during training.
For TAROT, PL, ARTUDA, and SRoUDA, the step size during training is defined as \( \frac{\varepsilon}{4 \times 255} \), with 10 steps per iteration. For model selection, we evaluate using PGD20 with \( \varepsilon \) and the same step size of \( \frac{\varepsilon}{4 \times 255} \), using a batch size of 32. Optimization is performed using SGD with a momentum of \( 0.9 \), a weight decay of \( 5 \times 10^{-4} \), and an initial learning rate of \( 0.005 \). These settings ensure consistency and robustness across all algorithms under evaluation.


\section{Additional Experimental Results}
\label{app:add_experiments}
Here, we present experimental results that were not included in the manuscript. 
Additionally, we perform supplementary experiments to further support the effectiveness of our proposed method, TAROT. 


\subsection{Essentially Domain-Invariant Robustness}
In \Cref{table:invariant-features}, we present partial performance results of PL and TAROT on the source and unseen domains on OfficeHome dataset, when $\varepsilon=8/255$. 
Here, we present the unreported values in \Cref{table:invariant-features_full}. 
In \Cref{table:invariant-features_full}, we observe that TAROT consistently outperforms its competitors in terms of robust accuracy, as shown in \Cref{table:invariant-features}. 
The only notable competitor in terms of standard accuracy is RFA. 
However, its robust accuracy is significantly lower than that of TAROT. 
Furthermore, TAROT outperforms other methods, across all metrics except ith only a few exceptions. 
In summary, TAROT demonstrates superior performance on both source and unseen domains compared to its competitors, owing to its ability to learn essentially domain-invariant robust features. 


\begin{table*}[!ht]
    \caption{\textbf{Performances of PL and TAROT on Source Domain and Unseen Domain, on OfficeHome.} Standard accuracy (\%) / Robust accuracy (\%) for AA with $\varepsilon=8/255$. Bold numbers indicate the best performance.
    }
    \scriptsize
    \centering
    \begin{tabular}{c|c|cc|c}
    \specialrule{1pt}{0pt}{0pt}
     & \multicolumn{1}{c|}{\textbf{Source}} & \multicolumn{2}{c|}{\textbf{Unseen}}  \\
    \specialrule{1pt}{0pt}{0pt}
    \textbf{Method} & Ar $\rightarrow$ Pr(Ar) & Ar $\rightarrow$ Pr(Cl) & Ar $\rightarrow$ Pr(Rw) & Avg. \\
    \hline
    ARTUDA & 62.59 / 8.53 &	29.46 / 11.02 & 32.29 / 7.30 &	41.45 /	8.95 \\
    RFA & \textbf{99.63} / 37.33 & 40.21 / 18.05 & \textbf{60.29} / 19.05 & \textbf{66.71} / 24.81 \\
    SRoUDA & 22.46 / 5.11 &	31.32 /	15.92 & 41.57 /	15.91 &	31.78 /	12.31 \\
    PL  & 24.68  / 10.88 & 35.51 / 24.72 & 43.84 / 25.25 & 34.68 / 20.28 \\
    TAROT     & 98.31 / \textbf{43.02} & \textbf{43.05} / \textbf{27.15} &	56.14 / \textbf{27.77} & 65.83 /	\textbf{32.65} \\
    \specialrule{.8pt}{0pt}{0pt}
     & Pr $\rightarrow$ Ar(Pr) & Pr $\rightarrow$ Ar(Cl) & Pr $\rightarrow$ Ar(Rw) & Avg. \\
    \hline
    ARTUDA & 66.16 / 23.00 & 20.87 / 7.45 &	24.08 /	5.90 & 37.04 / 12.12 \\
    RFA & \textbf{96.71} / 67.20 &	40.02 / 17.82 &	\textbf{59.79} / 19.37 &	\textbf{65.51} / 34.80 \\
    SRoUDA & 59.21 / 50.89 & \textbf{40.82} / 22.12 & 41.91 / 21.00 & 47.31 / 31.34 \\
    PL  &  45.21 / 27.01  & 33.47 / 22.09 & 44.60 / 23.07 & 41.09 / 24.05  \\
    TAROT     & 96.33 /	\textbf{78.69} &	40.12 /	\textbf{25.68} & 54.35 /	\textbf{28.21} & 63.60 / \textbf{44.19} \\
    \specialrule{1pt}{0pt}{0pt}
     & Cl $\rightarrow$ Rw(Cl) & Cl $\rightarrow$ Rw(Ar) & Cl $\rightarrow$ Rw(Pr) & Avg. \\
    \hline
    ARTUDA & 84.77 /55.81 &	14.34 /	3.21 &	27.26 /	13.99 &	42.12 /	24.34 \\
    RFA & \textbf{95.35} / 84.01 & \textbf{40.38} / 8.82 & 54.86 / 22.19 & \textbf{63.53} / 38.34 \\
    SRoUDA & 48.29 / 35.58 & 33.87 / 15.62 & 48.19 / 33.09 & 43.45 / 28.10 \\
    PL &  48.75 / 35.95 & 36.30 / 16.69 & 50.80 / 35.66 & 45.28 / 29.43 \\
    TAROT     &  93.65 / \textbf{84.35} & 39.72 / \textbf{17.18} & \textbf{55.46} / \textbf{37.67}  & 62.95 / \textbf{46.40} \\
    \specialrule{1pt}{0pt}{0pt}
     & Rw $\rightarrow$ Cl(Rw) & Rw $\rightarrow$ Cl(Ar) & Rw $\rightarrow$ Cl(Pr) & Avg. \\
    \hline
    ARTUDA & 85.40 / 22.08 & 31.64 / 5.15 &	51.59 / 18.72 &	56.21 / 15.32 \\
    RFA & \textbf{99.59} / 38.54 & \textbf{46.90} / 7.50 & \textbf{64.79} / 23.36 & \textbf{70.42} / 23.13 \\
    SRoUDA & 38.72 / 15.84 & 21.18 / 5.85 &	37.08 /	18.86 &	32.33 /	13.51 \\
    PL &  45.95 / 24.00 & 25.67 / 9.52 &	46.14 / 27.10 &  39.25 / 20.21 \\
    TAROT     &  97.68 / \textbf{51.55} & 42.73 / \textbf{11.83} &	63.39 /	\textbf{34.67} & 67.93 /	\textbf{32.68} \\
    \specialrule{1pt}{0pt}{0pt}
    & Ar $\rightarrow$ Cl(Ar) & Ar $\rightarrow$ Cl(Pr) & Ar $\rightarrow$ Cl(Rw) & Avg. \\
    \hline
    ARTUDA & 78.78 / 11.00 & 29.80 / 7.37 &	34.08 /	8.40 & 47.56 / 8.92 \\
    RFA & \textbf{99.63} / 33.79 & \textbf{49.83} / 17.59 & \textbf{60.50} / 17.37 & \textbf{69.99} / 22.92 \\
    SRoUDA & 30.70 / 7.87  & 34.47 / 15.68 & 35.85 / 13.54 & 33.67 / 12.36 \\
    PL &  32.51 / 11.83 &	38.30 /	\textbf{24.24} &	39.89 /	19.99 &	36.90 /	18.69 \\
    TAROT     & 99.59 /	\textbf{40.38} &	45.73 /	22.35 &	55.64 /	\textbf{22.42} &	66.98 /	\textbf{28.38} \\
    \specialrule{1pt}{0pt}{0pt}
    & Ar $\rightarrow$ Rw(Ar) & Ar $\rightarrow$ Rw(Cl) & Ar $\rightarrow$ Rw(Pr) & Avg. \\
    \hline
    ARTUDA & 18.83 / 2.64 & 7.70 / 0.89 & 6.28 / 0.45 & 10.94 / 1.33 \\
    RFA & \textbf{99.63} / 46.89 & 42.52 / 21.47 & 51.52 / 21.47 & 64.56 / 29.94 \\
    SRoUDA & 39.39 / 17.47 & 39.54 / 28.64 & 50.76 / 35.84 & 43.23 / 27.32 \\
    PL &  41.78 / 19.41 &	41.97 /	31.39 &	52.42 /	37.24 &	45.39 /	29.34 \\
    TAROT     & 98.35 /	\textbf{64.24} &	\textbf{47.86} /	\textbf{35.19} &	\textbf{58.32} /	\textbf{39.54} &	\textbf{68.18} /	\textbf{46.32}  \\
    \specialrule{1pt}{0pt}{0pt}
    & Cl $\rightarrow$ Ar(Cl) & Cl $\rightarrow$ Ar(Pr) & Cl $\rightarrow$ Ar(Rw) & Avg. \\
    \hline
    ARTUDA & 72.60 / 18.67 & 13.90 / 1.10 &	8.40 / 0.64 & 31.63 / 6.81 \\
    RFA & \textbf{95.37} / \textbf{84.81} & \textbf{48.61} / 19.49 & 50.52 / 20.11 & \textbf{65.84} / 42.47 \\
    SRoUDA & 39.86 / 25.98 & 33.61 / 18.14 & 39.55 / 21.23 & 37.67 / 21.78 \\
    PL &  42.45 / 28.94 & 37.33 / 22.39 & 42.92 / 24.81 & 40.90 / 25.38 \\
    TAROT     & 94.27 /	83.78 &	46.27 /	\textbf{27.96} &	\textbf{50.72} /	\textbf{26.76} &	63.76 /	\textbf{46.17} \\
    \specialrule{1pt}{0pt}{0pt}
    & Cl $\rightarrow$ Pr(Cl) & Cl $\rightarrow$ Pr(Ar) & Cl $\rightarrow$ Pr(Rw) & Avg. \\
    \hline
    ARTUDA & 72.99 / 38.10 & 13.14 / 3.05 &	19.92 /	5.90 & 35.35 / 15.68  \\
    RFA & \textbf{98.35} / 84.70 & \textbf{34.91} / 7.95 & \textbf{51.58} / 17.08 & \textbf{61.61} / 36.57 \\
    SRoUDA & 35.72 / 24.35 & 18.09 / 6.88 &	37.16 /	19.92 &	30.32 /	17.05 \\
    PL &  41.44 /	29.07 &	21.92 /	9.19 &	41.54 /	23.96 &	34.97 /	20.74 \\
    TAROT     &  95.65 	/ \textbf{86.09} &	30.20 / \textbf{11.54} &	51.41 / \textbf{27.34} &	59.09 /	\textbf{41.66} \\
    \specialrule{1pt}{0pt}{0pt}
    & Pr $\rightarrow$ Cl(Pr) & Pr $\rightarrow$ Cl(Ar) & Pr $\rightarrow$ Cl(Rw) & Avg. \\
    \hline
    ARTUDA & 97.30 / 47.94 & 23.28 / 3.79 & 45.12 / 11.59 & 55.23 / 21.11 \\
    RFA & \textbf{99.75} / 59.88 & \textbf{34.91} / 5.93 & \textbf{55.87} / 15.14 & \textbf{63.51} / 26.99 \\
    SRoUDA & 61.48 / 41.41 & 18.87 / 6.55 & 31.95 / 15.06 & 37.43 / 21.00 \\
    PL &   52.20 / 33.03 &	22.42 /	9.31 &	38.74 /	19.74 &	37.78 /	20.69  \\
    TAROT  & 98.60 /	\textbf{79.43} &	30.70 /	\textbf{9.48} &	53.27 /	\textbf{22.88} & 60.86 /	\textbf{37.26}  \\
    \specialrule{1pt}{0pt}{0pt}
    & Pr $\rightarrow$ Rw(Pr) & Pr $\rightarrow$ Rw(Ar) & Pr $\rightarrow$ Rw(Cl) & Avg. \\
    \hline
    ARTUDA & 98.29 / 60.17 & 25.67 / 4.62 &	36.24 /	16.24 &	53.40 / 27.01 \\
    RFA & \textbf{99.75} / 68.48 & 37.67 / 7.83 & 41.79 / 19.70 & 59.74 / 32.01 \\
    SRoUDA & 61.48 / 41.41 & 34.82 / 16.07 &	42.09 /	30.68 &	46.13 /	29.38 \\
    PL &   61.05 / 43.34 &	36.09 / \textbf{17.10} &	42.11 / 30.91 &	46.42 / 30.45 \\
    TAROT     & 96.13 /	\textbf{82.43} &	\textbf{39.14} / 16.15 &	\textbf{46.87} / \textbf{33.01} &	\textbf{60.71} / \textbf{43.86}  \\
    \specialrule{1pt}{0pt}{0pt}
    & Rw $\rightarrow$ Ar(Rw) & Rw $\rightarrow$ Ar(Cl) & Rw $\rightarrow$ Ar(Pr) & Avg. \\
    \hline
    ARTUDA & 90.66 / 19.72 & 41.19 / 15.46 &	53.17 / 16.99 &	61.67 / 17.39 \\
    RFA & \textbf{99.56} / 51.02 & \textbf{47.24} / 22.25 &	\textbf{64.09} / 27.33 &	\textbf{70.30} / 33.53 \\
    SRoUDA & 49.53 / 25.20 &	32.21 /	19.89 &	36.34 /	20.43 &	39.36 / 21.84 \\
    PL &  51.32 / 27.50 & 35.40 / 24.26 & 40.39 / 23.56 & 42.37 / 25.11 \\
    TAROT     & 95.50 /	\textbf{64.82} &	46.35 / \textbf{32.21} &	59.63 / \textbf{37.37} &	67.16 / \textbf{44.80}   \\
    \specialrule{1pt}{0pt}{0pt}
    & Rw $\rightarrow$ Pr(Rw) & Rw $\rightarrow$ Pr(Ar) & Rw $\rightarrow$ Pr(Cl) & Avg. \\
    \hline
    ARTUDA & 66.63 / 18.16 & 25.30 / 4.33 &  39.04 / 16.06 & 43.66 / 12.85 \\
    RFA & \textbf{99.59} / 41.52 & \textbf{48.26} / 7.99 & 44.35 / 20.18 & \textbf{64.07} / 23.23 \\
    SRoUDA & 44.27 / 21.57 & 20.03 /	6.84 & 33.49 /	20.89 &	32.60 /	16.44 \\
    PL &  48.80 / 26.60 & 23.65 / 9.60 & 36.98 / 24.86 & 36.47 / 20.35  \\
    TAROT & 97.89 / \textbf{61.85} &	42.23 / \textbf{14.30} &	\textbf{47.65} / \textbf{31.50} &	62.59 / \textbf{35.88} \\
    \specialrule{1pt}{0pt}{0pt}
    \end{tabular}
    \label{table:invariant-features_full}
\end{table*}

\subsection{Effect of Robust-PT on Various \texorpdfstring{$\varepsilon$}{Epsilon}}
\label{app:rob-PT}
We present the previously unreported values from \Cref{fig:robust_pt_effect} for the OfficeHome dataset. 
In \Cref{table:effect-robust-pt}, we provide the standard and robust accuracies of PL and TAROT across varying values of $\varepsilon$, both with and without Robust-PT. 
Notably, TAROT with Robust-PT consistently outperforms other methods. 
Is it worth emphasizing that Robust-PT is crucial for enhancing the performance of both PL and TAROT. 
As discussed in \Cref{ablation}, the performance gap between TAROT and PL widens as $\varepsilon$ increases.

\subsection{Sensitivity Analysis of \texorpdfstring{$\alpha$}{Alpha}}
\label{app:alpha}
We present the previously unreported values from \Cref{fig:sensitivity_alpha}. 
In \Cref{table:sensitivity_alpha_domainnet}, the results for DomainNet are reported. 
We can observe that the target performance is highest when $\alpha=1.0$. 
Additionally, $\alpha=1.0$ yields the best performance on both source and unseen (average) domains. 

In \Cref{table:sensitivity_alpha_visda}, the results for VisDA2017 results are reported. 
As shown in \Cref{fig:sensitivity_alpha}, the standard and robust accuracies on the target domain exhibit minimal variation across different values of $\alpha$. 
However, the performances on the source domain exhibit relatively large variations. 
We choose $\alpha=0.1$, since it shows highest robust accuracy among the candidate values of $\alpha$. 

\begin{table*}
    \caption{\textbf{Sensitivity Analysis of $\alpha$, on DomainNet.} Performance of generalization and robustness when $\alpha$ varies. 
    In each cell, the first number is the standard accuracy (\%), while the second number corresponds to the robust accuracy (\%) for AA. 
    }
    \footnotesize
    \centering
    \begin{tabular}{c|c|c|ccc|c}
    \specialrule{.8pt}{0pt}{0pt}
     & \multicolumn{1}{c|}{\textbf{Target}} & \multicolumn{1}{c|}{\textbf{Source}} & \multicolumn{4}{c}{\textbf{Unseen}}  \\
    \specialrule{.8pt}{0pt}{0pt}
    \textbf{$\alpha$} & C $\rightarrow$ R (R) & C $\rightarrow$ R (C) & C $\rightarrow$ R (I) & C $\rightarrow$ R (P) & C $\rightarrow$ R (S) & Avg. \\
    \hline
    0.0  & 43.57 / 28.68 &  48.41 / 35.89 & 10.70 / 5.66 & 24.44 / 11.46 &	22.84 / 13.55 & 29.99 / 19.05 \\
    0.05 & 46.24 / 30.66 & 52.79 / 39.19 & 26.28 / 12.19 & 26.28 / 12.19 & 24.80 / 14.68 & 32.24 / 20.50  \\
    0.1  & 46.83 / 31.03 & 55.87 / 41.88 & 26.65 / 12.49 & 26.65 / 12.49 & 26.08 / 15.59 & 33.42 / 21.41 \\
    0.5  & 49.39 / 31.46 & 67.71 / 51.57 & \textbf{30.18} / \textbf{13.57} & \textbf{30.18} / \textbf{13.57} & 34.07 / 19.81 & 38.93 / 24.59 \\
    1.0  & \textbf{49.73} / \textbf{31.73} & \textbf{71.58} / \textbf{54.42} & 14.36 / 6.60 & 31.45 / 13.53 & \textbf{36.26} / \textbf{20.29} & \textbf{40.68} / \textbf{25.32} \\
    \specialrule{.8pt}{0pt}{0pt}
    \end{tabular}
    \label{table:sensitivity_alpha_domainnet}
\end{table*}

\begin{table*}
    \caption{\textbf{Sensitivity Analysis of $\alpha$, on VisDA2017.} Performance of generalization and robustness when $\alpha$ varies. 
    In each cell, the first number is the standard accuracy (\%), while the second number corresponds to the robust accuracy (\%) for AA. 
    }
    \footnotesize
    \centering
    \begin{tabular}{c|c|c|c}
    \specialrule{.8pt}{0pt}{0pt}
     & \multicolumn{1}{c|}{\textbf{Target}} & \multicolumn{1}{c|}{\textbf{Source}}  \\
    \specialrule{.8pt}{0pt}{0pt}
    \textbf{$\alpha$} & Syn. $\rightarrow$ Real & Syn. &  Avg. \\
    \hline
    0.0  & \textbf{67.48} / \textbf{38.71} & 43.29 / 24.69 & 55.39 / 31.70
\\
    0.05 & 67.01 / 38.56 & 78.70 / 47.93 & 72.86 / 43.25\\
    0.1  & 66.12 / 37.91 & 85.18 / \textbf{51.21} & 75.65 / \textbf{44.56} \\
    0.5  & 66.45 / 36.97 & \textbf{86.63} / 46.30 & \textbf{76.54} / 41.64 \\
    1.0  & 64.48 / 35.48 & 67.63 / 34.32 & 66.06 / 34.90 \\
    \specialrule{.8pt}{0pt}{0pt}
    \end{tabular}
    \label{table:sensitivity_alpha_visda}
\end{table*}

\subsection{Evidence on Local Lipschitz Surrogate}
In constructing the objective for TAROT, we employ adversarial training to reduce the local Lipschitz constant. 
Here, we empirically demonstrate that combining adversarial training with pseudo labeling effectively reduces the local Lipschitz constant. Following the approach of \citet{yang2020closer}, we compute the empirical local Lipschitz constant using the following formula:
\small
\begin{equation}
\label{eq:local_lipschitz_empirical}
\frac{1}{n}\sum_{i=1}^n \max_{\bm{x}'_i \in \mathcal{B}_{\infty}(\bm{x}_i, \epsilon)} \dfrac{\lVert f(\bm{x}_i) - f(\bm{x}'_{i}) \rVert_1}{\lVert \bm{x}_i - \bm{x}'_i \rVert_\infty}
\end{equation}
\normalsize
Table \ref{table:lipschitz} illustrates the training dynamics of the local Lipschitz constants, showing that adversarial training with pseudo labels effectively reduces these constants during training phase of PL. 
We evaluate the empirical local Lipschitz constant under various settings, considering four cases: 
with or without Robust-PT, and with or without adversarial training. 
We observe that when conducting an adversarial training, the empirical local Lipschitz constant significantly decreases across all tasks. 

\begin{table*}[!t]
    \caption{\textbf{Empirical Local Lipschitz Constant in Various Training Settings.} \textbf{Lipschitz} denotes the empirical local Lipschitz constant value. Standard accuracy (\%) and the robust accuracy (\%) for PGD20 are also described. 
    }
    \footnotesize
    \centering
    \begin{tabular}{c|c|cc|cc|cc}
    \specialrule{.8pt}{0pt}{0pt}
    \textbf{Method} & Adv. Train. & \textbf{Lipschitz}  & Ar $\rightarrow$ Rw &  \textbf{Lipschitz}  & Cl $\rightarrow$ Rw & \textbf{Lipschitz}  & Pr $\rightarrow$ Rw\\
    \hline
    PL & \XSolidBrush & 6653.58 & \textbf{78.40} / 1.31  & 6518.85 & 72.09 / 2.50 & 6992.54 & 78.84 / 1.26 \\
    PL & $\varepsilon=8/255$ & \textbf{1014.95} & 77.78 / \textbf{70.53} & \textbf{981.91} & \textbf{72.80} / \textbf{64.66} & \textbf{1086.23} & \textbf{79.30} / \textbf{72.14}  \\
    \specialrule{.8pt}{0pt}{0pt}
    \end{tabular}
    \label{table:lipschitz}
\end{table*}

\begin{table*}
    \caption{\textbf{Effect of Robust-PT with various $\varepsilon$, on OfficeHome.} In each cell, the first number is the standard accuracy (\%), while the second number corresponds to the robust accuracy (\%) for AA. Bold numbers indicate the best performance.
    }
    \centering
    \footnotesize
    \begin{tabular}{c|c|c|cccc}
    \specialrule{1pt}{0pt}{0pt}
    $\varepsilon$ & Robust-PT &\textbf{Method} & Ar $\rightarrow$ Rw & Cl $\rightarrow$ Rw & Pr $\rightarrow$ Rw & Avg.\\
    \specialrule{1pt}{0pt}{0pt}
    \multirow{4}*{16/255} 
    & \Checkmark & PL & 73.10 / 40.26 & 68.14 / 37.37 & 74.82 / 40.74 & 72.02 / 39.45 \\
    & \XSolidBrush & PL & 6.59 /	0.00 & 3.121 / 0.00 & 5.92 / 0.161 & 5.21 / 0.05 \\
    \cline{3-7}
    & \Checkmark & TAROT &  \textbf{77.78} / \textbf{42.62} &	\textbf{71.31} / \textbf{39.22} &	\textbf{78.72} /	\textbf{43.13} & \textbf{75.94} / \textbf{41.66} \\
    & \XSolidBrush & TAROT & 22.47 / 0.74 &	21.71 /	0.90 & 18.98 / 0.73 & 21.05 / 0.79 \\
    \specialrule{1pt}{0pt}{0pt}
    \multirow{4}*{12/255} 
    & \Checkmark & PL & 78.15 / 55.68 &	71.70 /	50.06 &	78.29 /	54.17 &	76.05 /	53.30  \\
    & \XSolidBrush & PL & 8.54 / 0.05  &	5.30 / 0.00  &	5.92 / 0.34 & 6.59 / 0.13 \\
    \cline{3-7}
    & \Checkmark & TAROT &  \textbf{78.98} / \textbf{56.53} &	\textbf{72.39} / \textbf{52.15} &	\textbf{79.41} / \textbf{57.24} & \textbf{76.93} / \textbf{55.31}  \\
    & \XSolidBrush & TAROT & 23.04 / 1.81 &	37.53 / 3.83 & 28.80 / 0.62 & 29.79 / 2.09 \\
    \specialrule{1pt}{0pt}{0pt}
    \multirow{4}*{8/255} 
    & \Checkmark & PL &  78.70 /	69.43 &	72.53 /	63.44 &	78.27 / 69.50 & 76.50 /	67.45 \\
    & \XSolidBrush & PL & 10.83 / 1.68 & 12.65 / 2.32 & 9.00 / 1.17 & 10.83 / 1.72 \\
    \cline{3-7}
    & \Checkmark & TAROT & \textbf{78.77} / \textbf{70.46} & \textbf{73.01} / \textbf{63.78} & \textbf{79.44} / \textbf{70.83} & \textbf{77.07} / \textbf{68.36}  \\
    & \XSolidBrush & TAROT & 69.70 / 24.54 & 65.32 / 30.25 &	64.86 / 21.92 & 66.63 / 25.57 \\
    \specialrule{1pt}{0pt}{0pt}
    \multirow{4}*{6/255} 
    & \Checkmark & PL &  79.16 / 73.93 &	71.52 /	65.87 &	79.02 /	73.97 	& 76.57 / 71.26  \\
    & \XSolidBrush & PL & 63.39 / 31.86 & 59.24 / 29.49 & 51.85 / 20.82 & 58.16 / 27.39 \\
    \cline{3-7}
    & \Checkmark & TAROT &   \textbf{79.41}  / \textbf{74.36} & \textbf{72.48} / \textbf{67.27} & \textbf{79.57} / \textbf{75.24} & \textbf{77.16} / \textbf{72.29}  \\
    & \XSolidBrush & TAROT & 77.60 / 52.26 & 72.05 / 51.11 & 77.28 / 44.09 & 75.64 / 49.15 \\
    \specialrule{1pt}{0pt}{0pt}
    \multirow{4}*{4/255} 
    & \Checkmark & PL &  79.02 /	75.69 &	71.93  / 67.78 &78.59 / 75.56 & 76.51 / \textbf{74.75} \\
    & \XSolidBrush & PL & 78.31 / 60.13 & 71.22 / 55.91 & 77.83 / 58.05 & 75.79 / 58.03 \\
    \cline{3-7}
    & \Checkmark & TAROT & 78.86 / \textbf{76.43} & \textbf{73.03} / \textbf{69.59} & 78.84 / \textbf{75.86} & 76.91 / 73.96  \\
    & \XSolidBrush & TAROT & \textbf{79.41} / 69.20 & 72.53 / 62.57 & \textbf{79.94} / 70.16 	& \textbf{77.29} / 67.31 \\
    \specialrule{1pt}{0pt}{0pt}
    \multirow{4}*{2/255} 
    & \Checkmark & PL &  78.08 / 76.11 &	72.37 / 69.64 &	79.39 / 77.02 &	76.61 /	74.25 \\
    & \XSolidBrush & PL & 77.88 / 73.93 & 71.72 / 67.23 & 79.48 / 74.78  & 76.36 / 71.98 \\
    \cline{3-7}
    & \Checkmark & TAROT & 78.36 / \textbf{76.70} & \textbf{73.42} / \textbf{71.24} &	79.21 /	\textbf{77.09} & \textbf{77.00} /	\textbf{75.01}  \\
    & \XSolidBrush & TAROT & \textbf{78.56} / 72.24 & 72.37 / 68.65 & \textbf{79.62} / 75.65 & 76.85 / 72.18 \\
    \specialrule{1pt}{0pt}{0pt}
    \end{tabular}
    \label{table:effect-robust-pt}
\end{table*}

\subsection{Performance with Lower Perturbation Budgets \texorpdfstring{$\varepsilon$}{Epsilon}, on Office31 and OfficeHome.}
\label{app:lower_perturb}
We also conduct experiments with smaller values of $\varepsilon$ than those used in the main experiment in \cref{subsec:performance_target}. 
Specifically, we evaluate 
$\varepsilon \in \{8/255, 4/255\}$ on the Office31 and OfficeHome datasets, aligning with the experimental settings described in the original works \cite{lo2022exploring, awais2021adversarial, zhu2023srouda}. 
Tables \ref{table:office31_epsilon=8}, \ref{table:officehome_epsilon=8}, 
\ref{table:office31_epsilon=4} and \ref{table:officehome_epsilon=4}
demonstrate that TAROT also outperforms existing methods under small perturbation budgets. 
Compared to the other methods presented in Tables \ref{table:office31} and \ref{table:officehome}, which experience significant performance degradation at larger perturbation budgets ($\varepsilon=16/255$), TAROT maintains its robustness even under these larger perturbation budgets.

\subsection{Evaluation Against Other Attack Methods than AutoAttack}

We additionally evaluate TAROT and other existing methods against other attack methods than AutoAttack. 
We evaluate each methods on OfficeHome with perturbation size of $\varepsilon=16/255$, against FGSM, MM, CW20, PGD20 and AA.
In Table~\ref{table:other_attack_methods}, we can observe that TAROT outperforms existing methods in all means.

\begin{table*}[!t]
    \caption{\textbf{Performances of ARTUDA, RFA, SRoUDA, PL and TAROT on OfficeHome ($\varepsilon=16/255$), evaluated with FGSM, MM, CW20, PGD20 and AA.} Bold numbers indicate the best performance.}
    \footnotesize
    \centering
    \begin{tabular}{c|c|c|c|cccccc}
    \specialrule{1pt}{0pt}{0pt}
    \textbf{Method} & \textbf{Dataset} & \textbf{Task} & \textbf{Standard} & \textbf{FGSM} & \textbf{MM} & \textbf{CW20} & \textbf{PGD20} & \textbf{AA}\\
    \specialrule{1pt}{0pt}{0pt}
    ARTUDA & \multirow{5}{*}{OfficeHome} & All & 27.03 & 11.79 & 8.55 & 9.01 & 9.21  & 7.86 \\
    RFA &  & All & 55.00 & 20.50 & 9.81 & 15.39 & 16.15  & 8.49 \\
    SRoUDA  & & All & 57.97 & 46.64 & 36.21 & 41.49 & 42.56 & 33.42 \\
    PL &   & All & 66.00 & 55.08 & 47.38 & 51.08 & 51.71 & 44.38 \\
    TAROT &  & All & \textbf{68.29} & \textbf{57.29} & \textbf{49.58} & \textbf{53.44} & \textbf{54.01}  & \textbf{46.80}\\
    \specialrule{1pt}{0pt}{0pt}
    \end{tabular}
    \label{table:other_attack_methods}
    \vskip -0.05in
\end{table*}

\subsection{On the Use of the Standard Margin Risk of the Source Domain}
If replacing $\mathcal{R}_{\mathcal{S}}(f)$ with $\mathcal{R}^{\text{rob}}_{\mathcal{S}}(f)$
burdens the computation cost, requiring to generate adversarial examples.
Moreover, it would result in a looser bound in theoretical perspective ($\because \mathcal{R}_{\mathcal{S}}(f) \leq \mathcal{R}^{\text{rob}}_{\mathcal{S}}(f)$) , making it less desirable. To demonstrate the superiority of the proposed algorithm, we present empirical results obtained by replacing $\mathcal{R}_{\mathcal{S}}(f)$ with $\mathcal{R}^{\text{rob}}_{\mathcal{S}}(f)$.
As seen in the table below, TAROT with $\mathcal{R}_\mathcal{S}(f)$ shows higher performance in both standard and robust accuracies than TAROT with $\mathcal{R}_\mathcal{S}^\text{rob}(f)$. 
Hence, the use of $\mathcal{R}_\mathcal{S}(f)$ rather than $\mathcal{R}_\mathcal{S}^\text{rob}(f)$ is justified both theoretically (a tighter bound) and empirically. 

\begin{table*}[!ht]
    \caption{\textbf{Performance comparison when using the standard margin risk and the robust margin risk on the source domain.} Bold numbers indicate the best performance.}
    \footnotesize
    \centering
    \begin{tabular}{c|c|c|c|cccccc}
    \specialrule{1pt}{0pt}{0pt}
    \textbf{Method} & \textbf{Dataset} & \textbf{Task} & \textbf{Stand} & \textbf{AA} \\
    \specialrule{1pt}{0pt}{0pt}
    TAROT w/ $\mathcal{R}_\mathcal{S}^{(\rho)}(f)$ (Ours) & OfficeHome & All & \textbf{68.29} & \textbf{46.80} \\
    TAROT w/ $\mathcal{R}^{\text{rob},(\rho)}_\mathcal{S}(f)$ & OfficeHome & All & 67.63 & 44.24  \\
    \specialrule{1pt}{0pt}{0pt}
    \end{tabular}
    \label{table:standard_risk_comparison}
\end{table*}

\begin{table*}[!t]
    \caption{\textbf{Performances of ARTUDA, RFA, SRoUDA, PL and TAROT on Office31 ($\varepsilon=8/255$).} In each cell, the first number is the standard accuracy (\%), while the second number is the robust accuracy (\%) for AA. Bold numbers indicate the best performance.
    }
    \footnotesize
    \centering
    \begin{tabular}{c|cccccc|c}
    \specialrule{1pt}{0pt}{0pt}
    \textbf{Method} & A $\rightarrow$ D  & A $\rightarrow$ W & D $\rightarrow$ A & D $\rightarrow$ W & W $\rightarrow$ A & W $\rightarrow$ D & Avg.\\
    \specialrule{1pt}{0pt}{0pt}
    ARTUDA & 47.79 / 45.58 & 47.67 / 45.16 & 42.88 / 33.12 & 88.81 / 86.54 & 59.99 / 36.74 & 94.18 / 91.57 & 63.55 / 56.45 \\
    RFA & 78.51 / 45.18 & 73.84 / 33.08 & 62.30 / 46.57 & 98.24 / 79.87 & 61.02 / 43.95 &	99.20 /	81.53 & 78.85 / 55.03 \\
    SRoUDA & 89.96 / 85.54 	& 91.57 / 90.57 & 49.38 / 22.36 & 97.99 / 90.31 & 71.92 / 65.71 & 98.59 / 97.99 & 83.24 / 75.41  \\
    PL & \textbf{93.37}	/ \textbf{93.37} & \textbf{94.72} / 94.34 &	73.59 / 71.81 & 98.49 / 98.37 &	\textbf{74.26} / \textbf{72.63} &	99.80  / 99.60 & 89.04 / 88.35 \\
    \hline
    TAROT & \textbf{93.37} /	92.97 & 94.47 / \textbf{94.47} &	\textbf{76.32} / \textbf{75.19} &	\textbf{98.62} /	\textbf{98.49} &	72.74 /	71.64 &	\textbf{100.00} / \textbf{100.00} & \textbf{90.45} / \textbf{90.04} \\
    \specialrule{1pt}{0pt}{0pt}
    \end{tabular}
    \label{table:office31_epsilon=8}
\end{table*}
\begin{table*}[!t]
    \caption{\textbf{Performances of ARTUDA, RFA, SRoUDA, PL and TAROT on OfficeHome ($\varepsilon=8/255$).} In each cell, the first number is the standard accuracy (\%), while the second number is the robust accuracy (\%) for AA. Bold numbers indicate the best performance.
    }
    \footnotesize
    \centering
    \begin{tabular}{c|cccccc|c}
    \specialrule{1pt}{0pt}{0pt}
    \textbf{Method} & Ar $\rightarrow$ Cl  & Ar $\rightarrow$ Pr & Ar $\rightarrow$ Rw & Cl $\rightarrow$ Ar & Cl $\rightarrow$ Pr & Cl $\rightarrow$ Rw & \\
    \specialrule{1pt}{0pt}{0pt}
    ARTUDA & 47.45 / 32.33 	& 34.94 / 18.00 &	40.44 / 21.16 & 21.59 / 12.20 &	43.23 /	27.06 &	40.40 /	24.03 \\
    RFA & 47.49 / 31.59 & 53.80 / 29.13 & 62.98 / 28.44 & 43.55 / 16.32 	& 59.36 / 32.55 & 57.20 / 25.78 \\
    SRoUDA & 53.61 / 46.64 & 75.22 / 66.57 & 78.56 / 69.89 & 60.07 / 54.68 & 70.06 / 67.13 & 70.07 / 62.70 \\
    PL &  56.01 / 52.92 & 72.58 / 68.37 & 78.63 / 68.99 & 60.82 / \textbf{55.71} & \textbf{72.88} / \textbf{68.53} & 72.64 / 63.19 \\
    \hline
    TAROT & \textbf{56.58} / \textbf{53.28} & \textbf{75.36} / \textbf{71.50} &	\textbf{79.09} / \textbf{70.62} &	\textbf{61.06} / 55.30 &	72.52 / 68.17 &	\textbf{73.06} /	\textbf{63.92} \\
    \specialrule{1pt}{0pt}{0pt}
    & Pr $\rightarrow$ Ar  & Pr $\rightarrow$ Cl & Pr $\rightarrow$ Rw & Rw $\rightarrow$ Ar & Rw $\rightarrow$ Cl & Rw $\rightarrow$ Pr & Avg.\\
    \specialrule{1pt}{0pt}{0pt}
    ARTUDA  & 	27.73 / 9.81 & 46.76 / 37.39 & 49.46 / 28.02 & 32.18 / 17.18 & 54.85 / 43.71 & 68.12 / 40.03 & 42.26 / 25.91 \\
    RFA & 42.32 / 14.50 & 47.61 / 28.34 & 64.13 / 25.89 & 54.88 / 19.04 & 55.62 / 33.31 & 72.76 / 37.26 & 55.14 / 26.84 \\
    SRoUDA &  61.64 / \textbf{58.51} & 44.74 / 41.51 & 79.39 / \textbf{71.06} & \textbf{72.64} / \textbf{69.76} & 52.28 / 46.30 & 83.56 / 80.38 & 60.07 / 54.68 \\
    PL & 61.10 / 56.20 	& 52.81 / 49.71 & 78.63 / 69.06 & 72.60  / 67.74 &	60.21 / \textbf{56.63} & 84.14 / 80.42 &	68.59 / 63.12 \\
    \hline
    TAROT & \textbf{61.95} / 55.79 &	\textbf{54.09} / \textbf{50.84} &	\textbf{79.62} / 70.65 & 72.56 /	68.56 &	\textbf{60.28} / 55.67 &	\textbf{84.66} / \textbf{80.74} & \textbf{69.23} / \textbf{63.75} \\
    \specialrule{1pt}{0pt}{0pt}
    \end{tabular}
    \label{table:officehome_epsilon=8}
\end{table*}

\begin{table*}[!t]
    \caption{\textbf{Performances of ARTUDA, RFA, SRoUDA, PL and TAROT on Office31 ($\varepsilon=4/255$).} In each cell, the first number is the standard accuracy (\%), while the second number is the robust accuracy (\%) for AA. Bold numbers indicate the best performance.
    }
    \footnotesize
    \centering
    \begin{tabular}{c|cccccc|c}
    \specialrule{1pt}{0pt}{0pt}
    \textbf{Method} & A $\rightarrow$ D  & A $\rightarrow$ W & D $\rightarrow$ A & D $\rightarrow$ W & W $\rightarrow$ A & W $\rightarrow$ D & Avg.\\
    \specialrule{1pt}{0pt}{0pt}
    ARTUDA & 71.89	/ 71.69 	& 73.71 / 73.33 & 57.93 / 52.25 & 93.21 / 93.08 &	58.93 /	52.68 & 98.39 /	97.99 &	75.68 /	73.50 \\
    RFA & 83.53 / 78.11 & 81.89 / 72.58 & 61.38 / 54.03 & 97.48 / 96.73 &	63.44 / 56.12 &	\textbf{100.00} / 99.20 	& 81.29 / 76.13 \\
    SRoUDA &  92.97 / 92.77 & \textbf{95.22} / \textbf{94.21} & 74.62 / 65.74 & \textbf{98.74} / \textbf{98.74} & 66.45 / 64.57 & \textbf{100.00} / \textbf{100.00} & 88.00 / 86.01 \\
    PL & 89.56 / 89.56 & 93.46 	/ 93.33 & 75.04 / \textbf{74.55} & 98.49 / 98.49 &	72.70 / 72.70 &	\textbf{100.00}	/ \textbf{100.00} & 88.21 / 88.11 \\
    \hline
    TAROT & \textbf{93.37} / \textbf{93.17} & 93.84 / 93.59 &	\textbf{75.22} / \textbf{74.55} &	 98.49 / 98.49 & \textbf{74.51} /	\textbf{73.55} &	\textbf{100.00} / \textbf{100.00}	& \textbf{91.00} / \textbf{90.72} \\
    \specialrule{1pt}{0pt}{0pt}
    \end{tabular}
    \label{table:office31_epsilon=4}
\end{table*}
\begin{table*}[!t]
    \caption{\textbf{Performances of ARTUDA, RFA, SRoUDA, PL and TAROT on OfficeHome ($\varepsilon=4/255$).} In each cell, the first number is the standard accuracy (\%), while the second number is the robust accuracy (\%) for AA. Bold numbers indicate the best performance.
    }
    \footnotesize
    \centering
    \begin{tabular}{c|cccccc|c}
    \specialrule{1pt}{0pt}{0pt}
    \textbf{Method} & Ar $\rightarrow$ Cl  & Ar $\rightarrow$ Pr & Ar $\rightarrow$ Rw & Cl $\rightarrow$ Ar & Cl $\rightarrow$ Pr & Cl $\rightarrow$ Rw & \\
    \specialrule{1pt}{0pt}{0pt}
    ARTUDA & 49.44 	/	44.38 &	46.81 /	38.97 &	57.56 /	44.78 	& 38.53 / 31.23 	& 57.51 / 51.07 & 55.27 / 45.15 \\
    RFA & 49.21 / 40.18 & 58.80 / 45.28 & 69.20 / 48.73 & 50.23 / 29.30 & 63.11 / 48.46 & 62.80 / 42.46 \\
    SRoUDA & 55.44 / 51.84 & \textbf{76.48} / \textbf{74.34} & 79.00 / \textbf{77.19} & 61.27 / 58.96 & 68.06 / 66.91 & 69.38 / 67.32 \\
    PL & \textbf{55.79} / \textbf{54.18} & 75.29 / 72.74 & 78.01 / 75.08 & 61.72  / 59.54 & 71.91 / 69.27 & 72.16 / 69.02 \\
    \hline
    TAROT & 55.76 / 53.93 & 75.27 / 73.10 & \textbf{79.27} / 75.88 & \textbf{62.65} / \textbf{60.55} &  \textbf{72.51} 	/ \textbf{70.51} & \textbf{73.01} / \textbf{69.58}  \\
    \specialrule{1pt}{0pt}{0pt}
    & Pr $\rightarrow$ Ar  & Pr $\rightarrow$ Cl & Pr $\rightarrow$ Rw & Rw $\rightarrow$ Ar & Rw $\rightarrow$ Cl & Rw $\rightarrow$ Pr & Avg.\\
    \specialrule{1pt}{0pt}{0pt}
    ARTUDA  & 39.31 / 29.46 & 52.42 / 48.11 & 63.23 / 52.33 & 50.35 / 42.56 & 54.22 / 54.34 & 73.71 / 64.90 & 53.20 / 45.61 \\
    RFA & 48.04 / 28.88 & 49.51 / 39.08 & 70.07 / 47.74 & 59.37 / 37.78 & 56.52 / 45.06 & 75.22 / 57.90  & 61.21 / 43.49 \\
    SRoUDA & \textbf{61.60} / \textbf{59.15} & 48.14 / 46.51 & \textbf{80.39} / 76.08 & \textbf{73.79} / 70.76 & 57.84 / 55.21 & 83.16 / 81.38 & 67.88 / 65.47 \\
    PL &  58.14 / 57.93 & 52.33 / 50.75 & 79.34 / \textbf{76.59} & 72.64 / 70.13 & 59.59 / 57.73 & \textbf{83.74} / \textbf{82.07} & 68.39 / 66.25  \\
    \hline
    TAROT & 60.65 / 58.92 &	\textbf{53.08} / \textbf{51.34} &	79.78 / 76.43 & 73.05 / \textbf{71.74} &	\textbf{59.92} / \textbf{57.92} &	83.44 / 81.87   & \textbf{69.03} / \textbf{66.82} \\
    \specialrule{1pt}{0pt}{0pt}
    \end{tabular}
    \label{table:officehome_epsilon=4}
\end{table*}

\end{document}